\documentclass{article}

 \usepackage[preprint]{neurips_2026}

\usepackage[utf8]{inputenc} %
\usepackage[T1]{fontenc}    %
\usepackage{hyperref}       %
\usepackage{url}    
\usepackage{booktabs}       %
\usepackage{amsfonts}       %
\usepackage{nicefrac}       %
\usepackage{microtype}      %
\usepackage{xcolor}         %
\usepackage{enumitem}
\usepackage{colortbl}
\usepackage{soul}           %
\usepackage{float}          %
\usepackage{graphicx}       %
\usepackage{amsmath}        %
\usepackage{amssymb}        %
\usepackage{mathtools}      %
\usepackage{array}          %
\usepackage{multirow}       %
\usepackage{tikz}           %
\usepackage{algorithm}      %
\usepackage{algpseudocode}  %
\usepackage{subcaption}     %
\usepackage{diagbox}        %
\usepackage{tcolorbox}      %
\usepackage{afterpage}      %
\usepackage{scalerel,xparse}%
\usepackage{capt-of,etoolbox}%
\usepackage{wrapfig}        %

\usepackage{pifont}
\newcommand{\cmark}{\ding{51}}\newcommand{\xmark}{\ding{55}}
\newcolumntype{P}[1]{>{\centering\arraybackslash}p{#1}}
\newcolumntype{N}{>{\centering\arraybackslash\footnotesize}m{.5in}}
\newcolumntype{G}{>{\centering\arraybackslash}m{34pt}}

\usepackage{tikz}
\usepackage{wrapfig}
\usetikzlibrary{decorations.pathmorphing, patterns, arrows.meta, shapes.geometric, calc}

\definecolor{obj1}{HTML}{0072B2}      %
\definecolor{obj2}{HTML}{E69F00}      %
\definecolor{apparatus}{HTML}{2C2C2C} %
\definecolor{accent}{HTML}{009E73}    %

\tikzset{
  ball1/.style={circle, draw=obj1, fill=obj1!40, line width=1pt, minimum size=7mm, inner sep=0pt},
  ball2/.style={circle, draw=obj2, fill=obj2!40, line width=1pt, minimum size=7mm, inner sep=0pt},
  block1/.style={rectangle, draw=obj1, fill=obj1!40, line width=1pt, minimum width=7mm, minimum height=5mm, inner sep=0pt},
  block2/.style={rectangle, draw=obj2, fill=obj2!40, line width=1pt, minimum width=9mm, minimum height=7mm, inner sep=0pt},
  apparatus/.style={draw=apparatus, line width=1pt},
  ghost1/.style={circle, draw=obj1!#1, fill=obj1!#1, line width=0.5pt, minimum size=7mm, inner sep=0pt},
  ghost2/.style={circle, draw=obj2!#1, fill=obj2!#1, line width=0.5pt, minimum size=7mm, inner sep=0pt},
  ghostblock1/.style={rectangle, draw=obj1!#1, fill=obj1!#1, line width=0.5pt, minimum width=7mm, minimum height=5mm, inner sep=0pt},
  ghostblock2/.style={rectangle, draw=obj2!#1, fill=obj2!#1, line width=0.5pt, minimum width=9mm, minimum height=7mm, inner sep=0pt},
  motion/.style={->, >=Stealth, line width=0.8pt, draw=apparatus!70},
  ground/.style={apparatus, line width=1pt}
}

\tikzset{
  ball1/.style={circle, draw=obj1, fill=obj1!50, line width=0.8pt, minimum size=5mm, inner sep=0pt},
  ball2/.style={circle, draw=obj2, fill=obj2!50, line width=0.8pt, minimum size=5mm, inner sep=0pt},
  block1/.style={rectangle, draw=obj1, fill=obj1!50, line width=0.8pt, minimum width=5mm, minimum height=4mm, inner sep=0pt},
  block2/.style={rectangle, draw=obj2, fill=obj2!50, line width=0.8pt, minimum width=7mm, minimum height=5mm, inner sep=0pt},
  appline/.style={draw=apparatus, line width=0.8pt},
  ground/.style={draw=apparatus, line width=1pt},
  motion/.style={->, >=Stealth, line width=0.7pt, draw=apparatus!70}
}

\tikzset{
  ball1s/.style={circle, draw=obj1, fill=obj1!50, line width=0.7pt, minimum size=3.5mm, inner sep=0pt},
  ball2s/.style={circle, draw=obj2, fill=obj2!50, line width=0.7pt, minimum size=3.5mm, inner sep=0pt},
  block1s/.style={rectangle, draw=obj1, fill=obj1!50, line width=0.7pt, minimum width=3.5mm, minimum height=2.8mm, inner sep=0pt},
  block2s/.style={rectangle, draw=obj2, fill=obj2!50, line width=0.7pt, minimum width=4.5mm, minimum height=3.5mm, inner sep=0pt},
  blockNs/.style={rectangle, draw=apparatus!70, fill=apparatus!15, line width=0.6pt, minimum width=4mm, minimum height=3mm, inner sep=0pt},  %
  panelbg/.style={fill=apparatus!4, draw=apparatus!15, line width=0.4pt, rounded corners=2pt},
  pivot/.style={circle, fill=apparatus, draw=apparatus, minimum size=1.2mm, inner sep=0pt},
}

\usepackage{makecell}
\usepackage{graphicx}
\usepackage{rotating} %

\title{Principia: Relational Physics Tests for Video Models}

\author{
\hspace{-25pt}
Varun Varma Thozhiyoor$^{1,*}$\hspace{10pt}
Shivam Tripathi$^{1,*}$\hspace{10pt}
Venkatesh Babu Radhakrishnan$^{1}$\hspace{10pt}
Anand Bhattad$^{2}$\\[2pt]
$^{1}$Indian Institute of Science\hspace{20pt}
$^{2}$Johns Hopkins University\\[2pt]
\href{https://PrincipiaBench.github.io}{Project: https://PrincipiaBench.github.io}
}

\begin{document}

\maketitle

\begin{figure*}[h]
    \centering
    \vspace{-6mm}
    \includegraphics[width=0.98\textwidth]{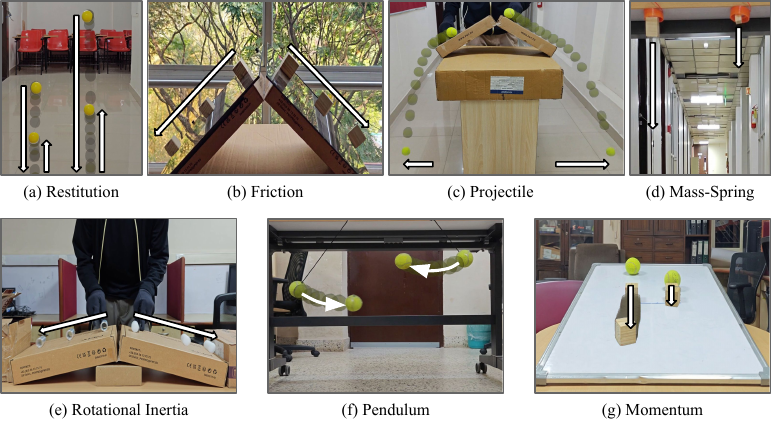}
    \vspace{-8pt}
\caption{\textbf{Principia} tests physical reasoning in video models through paired-object relational tests. 
Each panel shows a real-world experimental setup encoding a relational constraint between two objects obeying the same physical law (\textbf{a-g}); these constraints hold independent of camera, scale, or frame rate. We test whether modern video generators and vision-language models preserve them.}
    \label{fig:dataset_teaser}
\vspace{-2pt}
\end{figure*}

\begingroup
\renewcommand{\thefootnote}{}
\footnotetext{$^{*}$ Equal contribution.}
\endgroup

\begin{abstract}
Evaluating physical reasoning in video models is difficult because absolute motion measurements depend on frame rate, object scale, and camera calibration, all of which are often ambiguous or unavailable in generated video. We propose a different approach. When two objects in the same scene obey the same physical law, their motions must satisfy predictable relationships, and these relationships hold independent of calibration. We introduce \textit{Principia}, a benchmark that evaluates Newtonian physics through relational consistency between paired objects. Principia spans eight phenomena -- gravity, restitution, friction, rotational inertia, projectile motion, momentum, pendulum, and mass-spring oscillation—across translational, rotational, collisional, and oscillatory dynamics, using real-world scenes recorded under controlled protocols. We also introduce a calibration-independent consistency score that quantifies physical violation directly in image space. Across thousands of generations from six state-of-the-art video generators, no model exceeds 0.42 on Principia despite all scoring around 0.8 on VBench. Vision-language models are evaluated on their ability to detect relational physics violations, with the best model achieving only 67\% accuracy and most performing near chance level.
\end{abstract}
\begin{figure}[t!]
    \centering
   \includegraphics[width=\textwidth]{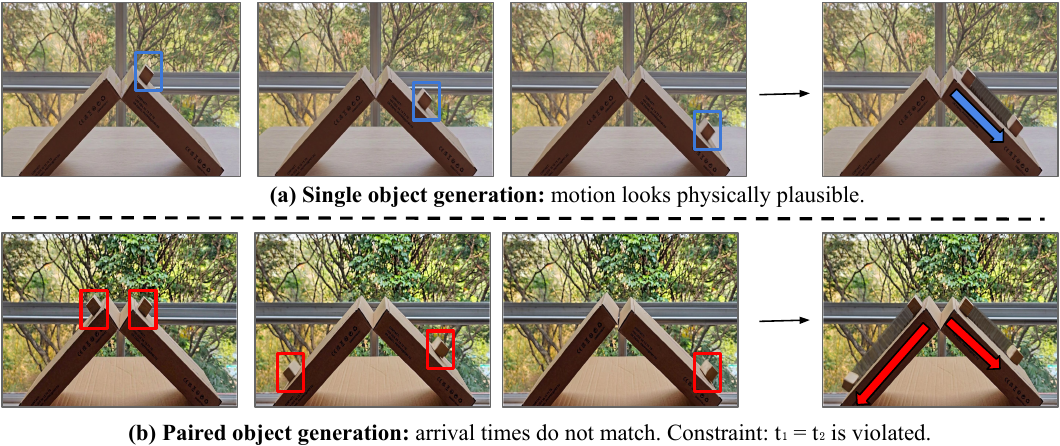}
    \vspace{-10pt}
\caption{\textbf{(a)} A generated video of a single block sliding down an incline, shown at three timesteps and as a stroboscopic composite (top right): motion appears plausible. 
\textbf{(b)} A generated video of two blocks with different masses released simultaneously on identical inclines, where the blocks arrive at different times, violating the mass-independence invariant of gravitational acceleration. This violation only becomes detectable through relative comparison. 
}
\label{fig:teaser2}
\vspace{-10pt}
 \end{figure}
\section{Introduction}
 Evaluation and benchmarks have been a strong basis for analyzing and improving modern machine learning models. Today’s video generators have reached a point where they can produce highly realistic clips that are hard to distinguish from real videos. VBench, one of the most widely used benchmarks for video generation, evaluates visual quality along several axes~\cite{huang2023vbench}. These have helped drive rapid progress in the field. However, in this paper, we show that visual realism does not imply physical consistency. 

To this end, several benchmarks have recently been proposed to evaluate physical consistency in generated videos. Some of these evaluate whether motion appears physically plausible~\cite{bansal2024videophyevaluatingphysicalcommonsense,bansal2025videophy}. Others compare generated trajectories against recorded videos~\cite{physicsiq,li2025pisa}. A few also evaluate motion against explicit physical laws~\cite {zhang2025morpheusbenchmarkingphysicalreasoning}. 
However, plausibility is subjective and can be met by visually convincing but physically incorrect motion. Moreover, trajectory matching penalizes deviation from a single reference even when multiple futures are physically valid, and law-based evaluation often requires recovering or assuming metric quantities. Single-phenomenon studies like free-fall under gravity~\cite{thozhiyoor2025objects} further reveal targeted failure modes but do not establish whether models preserve physical structure across different physics phenomena.

\looseness=-1 To make physical evaluation central to modern video generators, especially given their potential use as \emph{world models} for predicting the consequences of actions~\cite{craik1967nature,ha2018world}, we propose Principia, a physics evaluation benchmark that takes into account the relative motion of two objects and measures the extent to which their relative motions satisfy the expected physical relation for eight phenomena (Fig.~\ref{fig:dataset_teaser}). We evaluate the model's understanding of gravity in free fall, restitution measured from motion during and after collisions, sliding on inclined planes under friction, rotational inertia and momentum, projectile motion, pendulum swinging under gravity, and spring-mass systems under gravity.

The motivation for using relative motion is easily explained with the following example. Let us take a block moving down an inclined plane. In the case of a single block, as the visual input provides no direct information about material properties or mass, evaluation of the dynamics of sliding can only be carried out qualitatively, such as whether the block is accelerating down the inclined plane. 

Consider two blocks of different masses placed under identical conditions (Fig.~\ref{fig:teaser2}), i.e., having the same starting point on inclined planes with the same angle of inclination and coefficient of friction. Assuming that the two blocks follow the same physical laws, they will reach the bottom of the ramp simultaneously because the acceleration of a block sliding down an inclined plane with friction does not depend on its mass. This equality is an invariant of the underlying physical law; it should not depend on the absolute values of the physical variables. In general, physical laws impose invariants on the motion of multiple objects that can include equalities, ratios, and relative ordering in time and space. Such invariants can be computed from image-space measurements without estimating absolute quantities such as mass, scale, velocity, or acceleration.

For each physical phenomenon, we specify the expected relation between the motions of paired objects and measure how closely the generated motion satisfies that relation. We aggregate these deviations into a continuous \emph{invariance score}. The test is straightforward: if two objects are released under the same conditions, they must reach the ground at the same time. A systematic deviation from this relationship indicates a violation of the tested physical law, irrespective of whether either individual motion appears plausible.

However, creating such a benchmark is difficult because small geometric, temporal, surface, or release asymmetries may produce signals indistinguishable from genuine physics violations, requiring strict experimental control. We captured hundreds of real-world videos, ensured synchronized release and controlled experiments, and conducted automated and manual validation at all steps to filter out bad videos. The obtained benchmark includes about 500 scenes of eight Newtonian phenomena covering the translational, rotational, collisional, and oscillatory dynamics categories. 

The same relational tests can also be applied to VLMs. We additionally build a synthetic testbed in Isaac Sim\cite{NVIDIA_Isaac_Sim} to generate controlled and counterfactual scenarios at scale. This provides a scalable way to evaluate both video generators and VLMs, and can further be used to develop methods for improving their physical consistency.

\textbf{Contributions.}
\vspace{-5pt}
\begin{itemize}[leftmargin=*, itemsep=0pt]
\item \textbf{Principia}, a benchmark consisting of 500+ real-world scenes capturing 8 physics principles, with calibrated objects, aligned geometry, and coordinated release to facilitate comparison against analytic expectations.

\item An invariant \emph{score of relational consistency} which evaluates consistency relations purely in image space without any dependence on metric scale, velocity, or camera intrinsics.

\item A scalable synthetic pipeline implemented in Isaac Sim that generates physical scenarios, both controlled and counterfactual, allowing for evaluation of VLMs and video generators.

\item A benchmark evaluating six state-of-the-art video generators and four vision-language models reveals that no model achieves a score higher than $0.42$ on Principia despite achieving scores around 0.8 on standard visual benchmarks, and that increasing the size of a network does not improve its physical consistency.
\end{itemize}

\section{Related Work}

\noindent\textbf{Video generative models, world simulators, and physics-guided generation.}
Recent video diffusion and autoregressive models~\cite{ho2022video,micheli2023transformers,bruce2024genie,wiedemer2025video} achieve high visual fidelity and are increasingly described as world models~\cite{craik1967nature}. Probing studies suggest that such models somewhat encode 3D structure and motion~\cite{bhattad2023stylegan, bhattad2024stylitgan, du2023generative, xing2025luminet, elbanani2024probing,zhan2024general,bhattad2025visual,joseph2026interpreting,sarkar2024shadows}, but perceptual realism does not imply adherence to physical laws. Several approaches attempt to improve physical realism  through reward-based fine-tuning~\cite{li2025pisa,le2025newtonrewards}, trajectory correction~\cite{zhang2025videorepa}, or integration of physics engines~\cite{liu2024physgen,Yuan_2025_NewtonGen,wang2025physctrl,chen2025physgen3d}, all of which depend on how physical correctness is evaluated.

\noindent\textbf{Benchmarking physical reasoning.}
Existing benchmarks fall into four families (Table~\ref{tab:related_work}): \textit{simulation-based reasoning} in calibrated environments~\cite{riochet2018intphys,bordes2025intphys,bear2021physion,tung2023physion++,bakhtin2019phyre,li2024iphyre}; \textit{VQA-based VLM evaluation}~\cite{chow2025physbench,foss2025causalvqa,mvpbench}; \textit{generator commonsense scoring} via human or VLM judges~\cite{bansal2024videophyevaluatingphysicalcommonsense,bansal2025videophy,guo2025t2vphysbenchfirstprinciplesbenchmarkphysical,PhysGenBench,gu2025phyworldbenchcomprehensiveevaluationphysical}; and \textit{real-video physics evaluation}~\cite{physicsiq,zhang2025morpheusbenchmarkingphysicalreasoning,zhan2025inferring,garcia2025learning,li2025pisa}. All require absolute measurements—calibration, scale, or recorded ground-truth parameters—that may be ambiguous in generated video. The closest precursor is the two-object gravity protocol of Thozhiyoor et al.~\cite{thozhiyoor2025objects}, which sidesteps calibration by testing ratios of fall times to isolate Galileo's principle: unit-free, relational, and quantitative. Principia generalizes this single-phenomenon protocol to eight Newtonian phenomena spanning translational, rotational, collisional, and oscillatory dynamics, and evaluates both video generators and vision-language models.

\begin{table}[t!]
\centering
\caption{
\textbf{Comparison of physics benchmarks.}
Principia uniquely combines calibration-independent relational evaluation, quantitative physics-derived scoring, multi-phenomenon coverage, real video data, and evaluation of both video generators and vision-language models.
\textbf{Unit-Free}: evaluation independent of metric scale or camera calibration.
\textbf{Relational}: explicit intra-scene invariants between paired objects under the same physical law.
\textbf{Quant.\ Physics}: continuous score derived from the physical law being tested.
\textbf{Multi-Phys.}: spans multiple distinct physical phenomena.
\textbf{Real}: includes real-world video data.
\textbf{Gen}: evaluates video generators.
\textbf{VLM}: evaluates vision-language models.
}
\label{tab:related_work}
\setlength{\tabcolsep}{4pt}
\renewcommand{\arraystretch}{1.15}
\resizebox{\linewidth}{!}{%
\begin{tabular}{l ccc cc cc}
\toprule
\textbf{Work} 
& \textbf{Unit-Free} 
& \textbf{Relational} 
& \textbf{Quant.\ Physics} 
& \textbf{Multi-Phys.} 
& \textbf{Real} 
& \textbf{Gen} 
& \textbf{VLM} \\
\midrule
\multicolumn{8}{l}{\textit{\textcolor{gray!90}{Simulation-based reasoning}}} \\
IntPhys~\cite{riochet2018intphys,bordes2025intphys} 
& \xmark & \xmark & \xmark & \cmark & \xmark & \xmark & \cmark \\
PHYRE/IPHYRE~\cite{bakhtin2019phyre,li2024iphyre}, Physion~\cite{bear2021physion,tung2023physion++} 
& \xmark & \xmark & \xmark & \cmark & \xmark & \xmark & \xmark \\
\midrule
\multicolumn{8}{l}{\textit{\textcolor{gray!70}{VQA-based VLM evaluation}}} \\
PhysBench~\cite{chow2025physbench}, CausalVQA~\cite{foss2025causalvqa}, MVPBench~\cite{mvpbench} 
& \xmark & \xmark & \xmark & \cmark & \cmark & \xmark & \cmark \\
\midrule
\multicolumn{8}{l}{\textit{\textcolor{gray!70}{Generator commonsense \& plausibility}}} \\
VideoPhy~\cite{bansal2024videophyevaluatingphysicalcommonsense,bansal2025videophy}, T2VPhysBench~\cite{guo2025t2vphysbenchfirstprinciplesbenchmarkphysical}, 
& \xmark & \xmark & \xmark & \cmark & \xmark & \cmark & \xmark \\
PhyGenBench~\cite{PhysGenBench}, PhyWorldBench~\cite{gu2025phyworldbenchcomprehensiveevaluationphysical} & & & & & & & \\
NewtonRewards~\cite{le2025newtonrewards} 
& \xmark & \xmark & \cmark & \cmark & \xmark & \cmark & \xmark \\
\midrule
\multicolumn{8}{l}{\textit{\textcolor{gray!70}{Real-video physics evaluation}}} \\
Physics-IQ~\cite{physicsiq} 
& \xmark & \xmark & \xmark & \cmark & \cmark & \cmark & \xmark \\
Morpheus~\cite{zhang2025morpheusbenchmarkingphysicalreasoning} 
& \xmark & \xmark & \cmark & \cmark & \cmark & \cmark & \xmark \\
PhysVid~\cite{zhan2025inferring} 
& \xmark & \xmark & \cmark & \cmark & \cmark & \xmark & \cmark \\
Delfys75~\cite{garcia2025learning} 
& \xmark & \xmark & \cmark & \cmark & \cmark & \xmark & \xmark \\
PISA~\cite{li2025pisa} 
& \xmark & \xmark & \xmark & \xmark & \cmark & \cmark & \xmark \\
\midrule
\multicolumn{8}{l}{\textit{\textcolor{gray!70}{Unit-free relational}}} \\
Galileo's equivalence~\cite{thozhiyoor2025objects} 
& \cmark & \cmark & \cmark & \xmark & \xmark & \cmark & \xmark \\
\rowcolor{blue!15}
\textbf{Principia (ours)} 
& \textbf{\cmark} 
& \textbf{\cmark} 
& \textbf{\cmark} 
& \textbf{\cmark} 
& \textbf{\cmark} 
& \textbf{\cmark} 
& \textbf{\cmark} \\
\bottomrule
\end{tabular}
}
\vspace{-15pt}
\end{table}

\section{The Principia Dataset}
Principia consists of 500+ real-world paired-object scenes spanning eight Newtonian phenomena (Table~\ref{tab:rel_ratio}). Each scene contains two objects whose motions must satisfy a relational invariant independent of camera, scale, or frame rate. All parameters are held fixed except the one varied (mass, height, length, or moment of inertia), isolating the law being tested.
\subsection{Construction Protocol}

Relational invariants are sensitive to small perturbations: a slight asymmetry in incline angle, a timing skew between paired releases, or a small lateral push on a pendulum can produce a relational signal indistinguishable from a real physics violation. Construction error must be small enough that it does not obscure the violations we aim to detect, which makes data collection the bulk of the work. We recorded approximately 750+ videos in total and applied heavy filtering to retain those meeting the conditions each invariant assumes. The resulting dataset is further augmented using editing models\cite{google2026gemini31flashimage} to increase diversity, yielding a final dataset comprising 529 scenes.

We enforce four constraints during data collection. \textit{Matched geometry:} paired objects share identical contact surfaces, ramps, supports, or springs, machined or assembled to a common specification (for example, the rotational-inertia experiment uses a solid Delrin cylinder and a hollow aluminum cylinder manufactured in-house to match in mass, height, and outer radius within $5\%$). \textit{Synchronized release:} mechanical alignment guides hold both objects in matched starting poses and release them simultaneously; hand-released objects are avoided except for pendulum swings. \textit{Controlled surface material:} contact surfaces are validated for each pairing, with no-slip conditions inspected for rolling experiments and contact faces inspected for friction experiments. \textit{Minimized external forces:} lateral velocity at release is held below visible drift, air resistance is negligible on the relevant timescales, and phenomenon-specific assumptions are verified individually (no visible slip on rolling ramps, collision axis aligned with the motion direction for momentum).

We segment and track all objects with SAM3~\cite{carion2025sam3} using hand-annotated initial points. All subsequent measurements happen in pixel space; the relational consistency score depends only on ratios and equalities of pixel-space quantities, so evaluation requires no knowledge of camera intrinsics, frame rate, or metric scale.

Each session yields multiple takes per scene; we include only those passing manual inspection of the recorded video and the resulting SAM3 trajectory (for example, confirming monotonic descent for ramp scenarios, clean ball-on-ground transitions for restitution, and periodic motion for pendulum and mass-spring scenes). Common excluded failure modes include lateral drift, asymmetric release timing, pendulum motion outside the small-angle regime, ball spin biasing projectile trajectories, and surface anomalies producing slip in rolling experiments.

\subsection{Principia-Synth}
In addition to the recorded real-world videos, we construct a synthetic dataset for each phenomenon using Isaac Sim~\cite{NVIDIA_Isaac_Sim, prabhudesai2026solvingphysicsolympiadreinforcement}. The physics simulator allows us to render both physically correct videos and corresponding \emph{anti-physics} videos, which are generated by explicitly violating the relational invariant associated with each phenomenon. We use this Principia-Synth dataset to evaluate whether VLMs can identify relational physics failures. Since such anti-physics videos cannot be recorded in the real world, and AI-generated videos that violate physical constraints often contain visual artifacts that could confound the evaluation, we instead rely on simulation to generate controlled anti-physics scenarios. Representative examples of the real-world physics and anti-physics videos for each phenomenon are provided in section \ref{scenarios} and Appendix~\ref{app:principia-synth-anti-physics} respectively.

\subsection{Phenomena}
\label{scenarios}
Principia covers four types of Newtonian dynamics: \textit{translational}, \textit{rotational}, \textit{collisional}, and \textit{oscillatory} dynamics. Free-fall and projectile dynamics examine \textit{translational} dynamics under gravity, while inclined-plane dynamics additionally involve friction. The study of rotational inertia involves \textit{rotational} dynamics and the interaction between translational and rotational motion under no-slip rolling conditions. Restitution and momentum dynamics investigate \textit{collisional} dynamics and the relationships between the motions of colliding bodies. Pendulum and mass-spring dynamics investigate \textit{oscillatory} dynamics under gravity and elastic restoring forces. Representative samples from Principia-Synth are shown alongside the description of each phenomenon.

\begin{minipage}[c]{0.75\linewidth}
\paragraph{Restitution and Gravity.}
Two identical balls are dropped from different heights onto the same surface. The rebound height satisfies $r = e^2 h$, so the relational invariant is $r_1/h_1 = r_2/h_2$. The same recordings yield Galileo's gravity invariant $h_1/h_2 = (t_1/t_2)^2$.
\end{minipage}
\hfill
\begin{minipage}[c]{0.21\linewidth}
\centering
\includegraphics[
    width=\linewidth,
    trim=180px 50px 180px 130px,
    clip
]{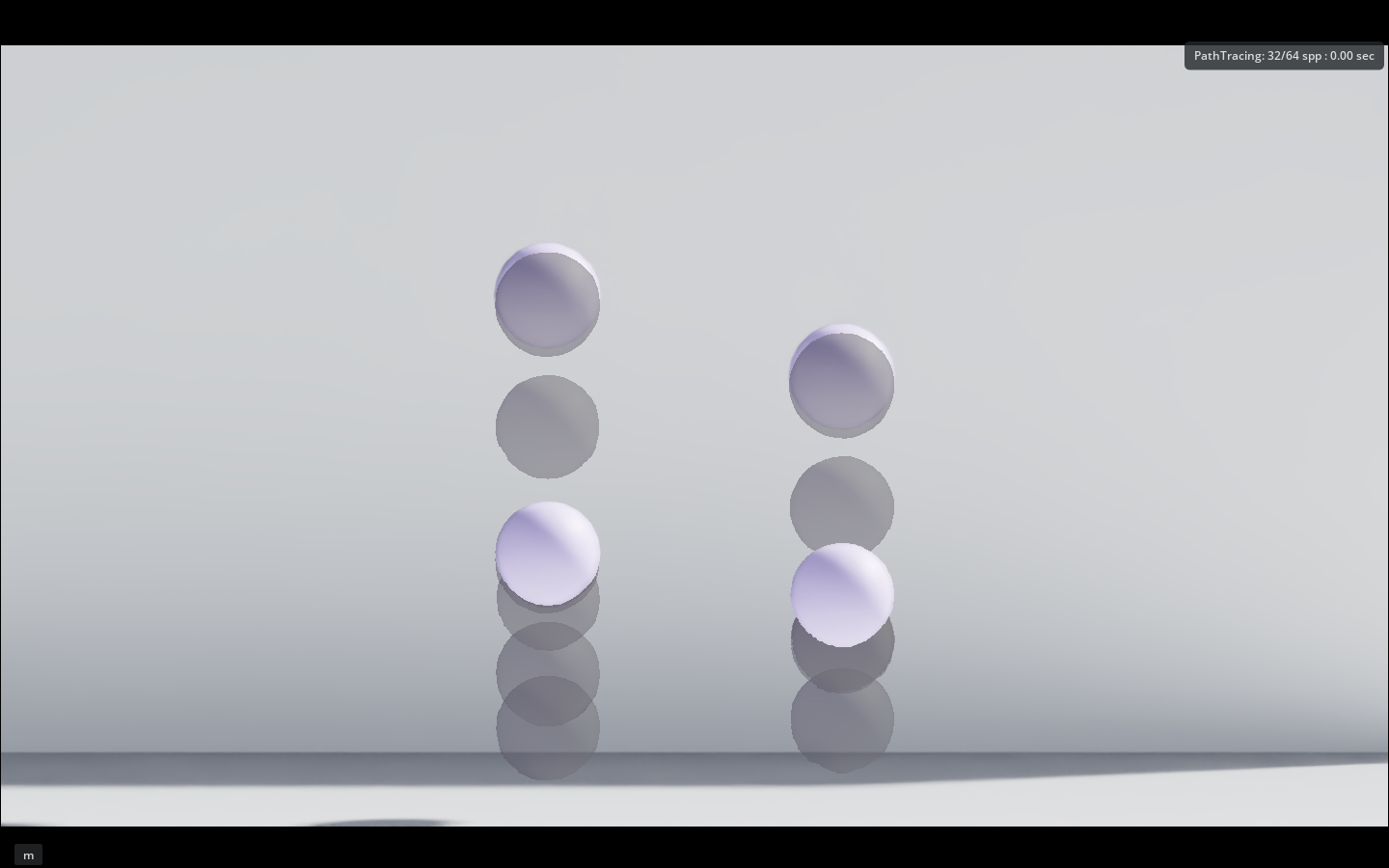}
\end{minipage}

\begin{minipage}[t]{0.75\linewidth}
\paragraph{Friction.}
\vspace{-30pt}
Two blocks of different mass slide down opposite faces of a tent-shaped ramp with identical surface material and angle. Because acceleration on an incline with friction satisfies $a = g(\sin\theta - \mu\cos\theta)$, the result is mass-independent. The relational invariant is therefore equality of arrival times, $t_1 = t_2$, despite the mass difference between the two blocks.
\end{minipage}
\hfill
\begin{minipage}[c]{0.21\linewidth}
\centering
\includegraphics[
    width=\linewidth,
    trim=0px 50px 0px 130px,
    clip
]{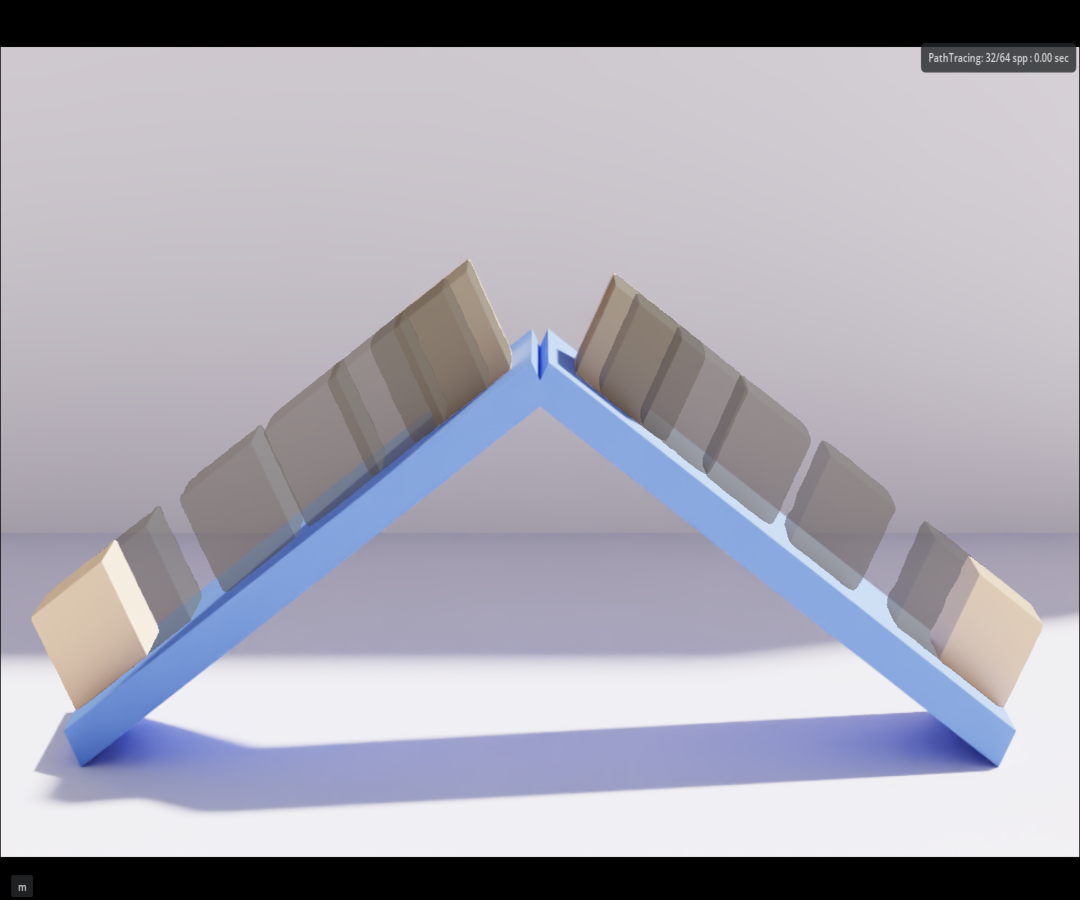}
\end{minipage}

\begin{minipage}[t]{0.75\linewidth}
\paragraph{Rotational Inertia.}
\vspace{-30pt}
A solid and a hollow cylinder of matched mass and radius roll down opposite faces of a no-slip ramp. Pure rolling acceleration is $a = g\sin\theta / (1 + I/mR^2)$, so the cylinder with greater moment of inertia arrives later. The relational invariant is the time ratio $t_1/t_2 = \sqrt{(1 + I_1/mR^2)/(1 + I_2/mR^2)}$.
\end{minipage}
\hfill
\begin{minipage}[c]{0.21\linewidth}
\centering
\includegraphics[
    width=\linewidth,
    trim=0px 50px 0px 130px,
    clip
]{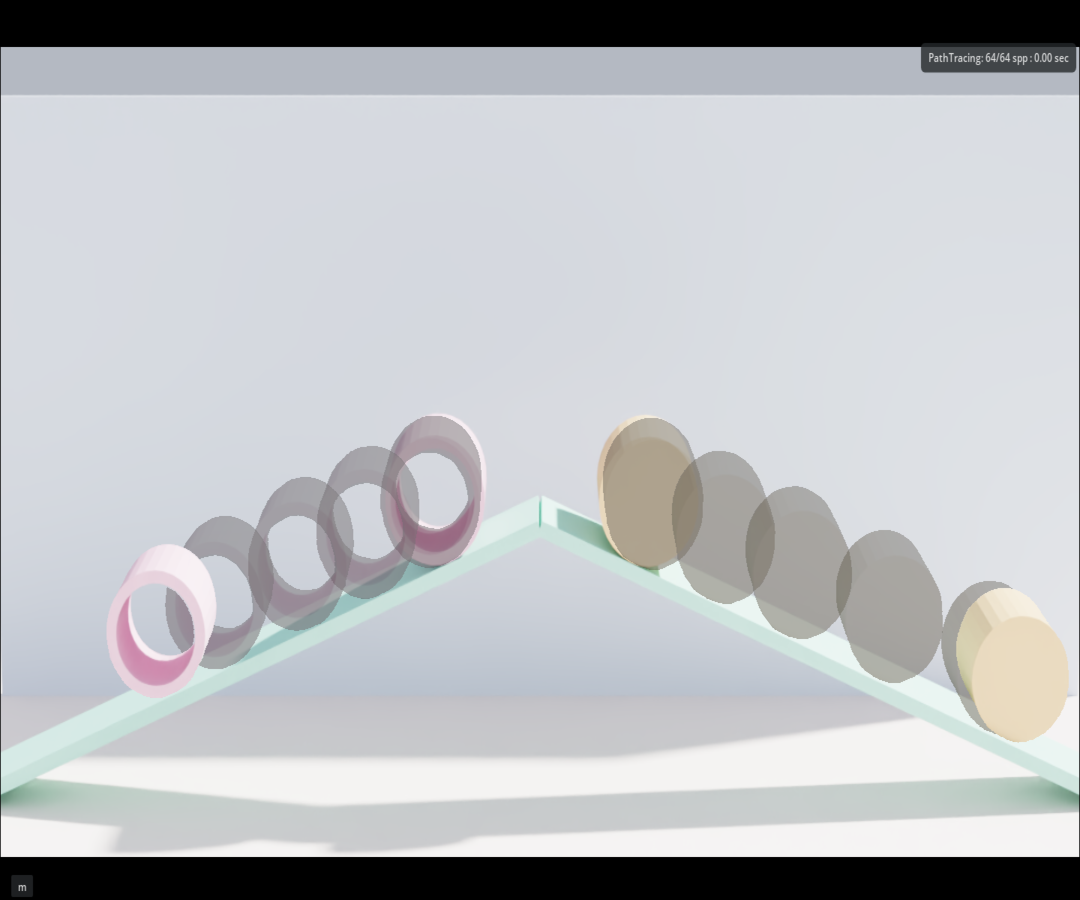}
\end{minipage}

\begin{minipage}[t]{0.75\linewidth}
\vspace{-30pt}
\paragraph{Momentum.}
Two identical balls are released from different heights on a single ramp and collide with identical blocks placed at matched distances. The ball released higher imparts greater momentum and pushes its block farther. The relational invariant is ($x \propto h$). In a second setup, identical balls collide with blocks of different masses. The heavier block travels a shorter distance, giving the relational invariant ($x \propto 1/m$).
\end{minipage}\hfill
\begin{minipage}[c]{0.21\linewidth}
\centering
\includegraphics[
    width=\linewidth,
    trim=0px 100px 0px 80px,
    clip
]{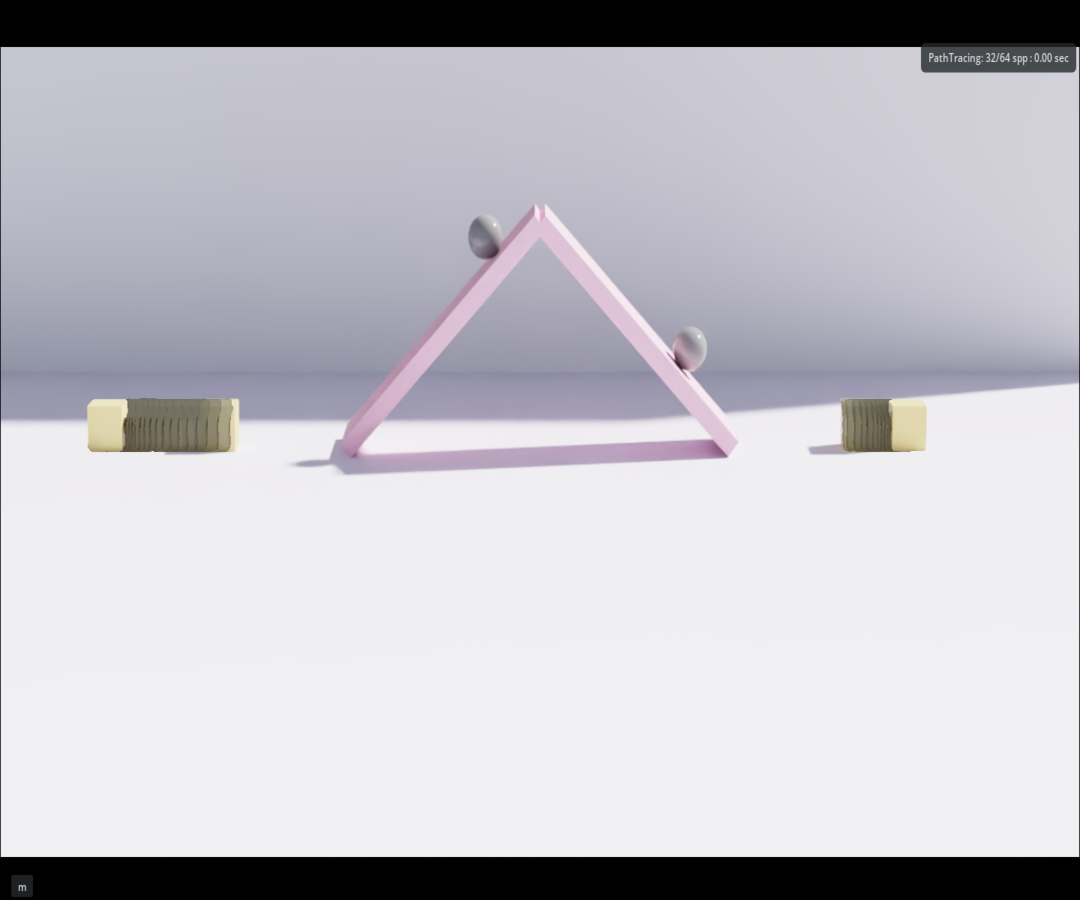}
\end{minipage}

\begin{minipage}[t]{0.74\linewidth}
\paragraph{Projectile Motion.}
\vspace{-30pt}
Two spheres are launched in opposite directions from a central platform at distinct heights with fixed launch angle. Greater height produces greater launch velocity ($v \propto \sqrt{h}$) and therefore greater range ($R \propto v$). The relational invariant is range ordering: the higher-launched sphere lands farther.
\end{minipage}\hfill
\begin{minipage}[c]{0.21\linewidth}
\centering
\includegraphics[
    width=\linewidth,
    trim=180px 50px 180px 130px,
    clip
]{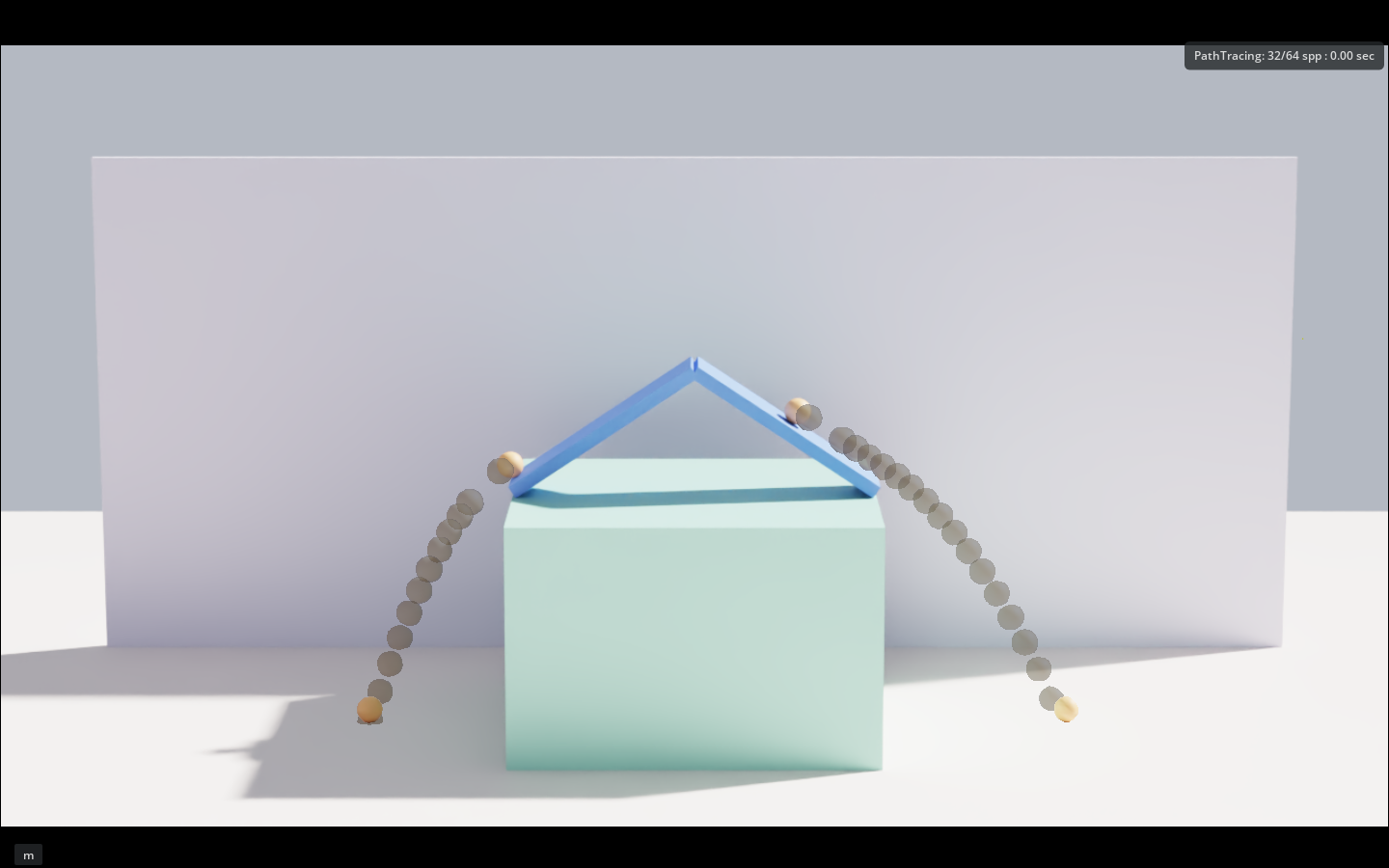}
\end{minipage}

\begin{minipage}[t]{0.75\linewidth}
\paragraph{Pendulum.}
\vspace{-30pt}
Two pendulums of different length and identical bob mass are released from the same small angle. The period $T = 2\pi\sqrt{l/g}$ is independent of bob mass, so the relational invariant is $T_1/T_2 = \sqrt{l_1/l_2}$. The longer pendulum therefore oscillates more slowly, with a proportionally larger period.
\end{minipage}\hfill
\begin{minipage}[c]{0.21\linewidth}
\centering
\includegraphics[
    width=\linewidth,
    trim=0px 50px 0px 130px,
    clip
]{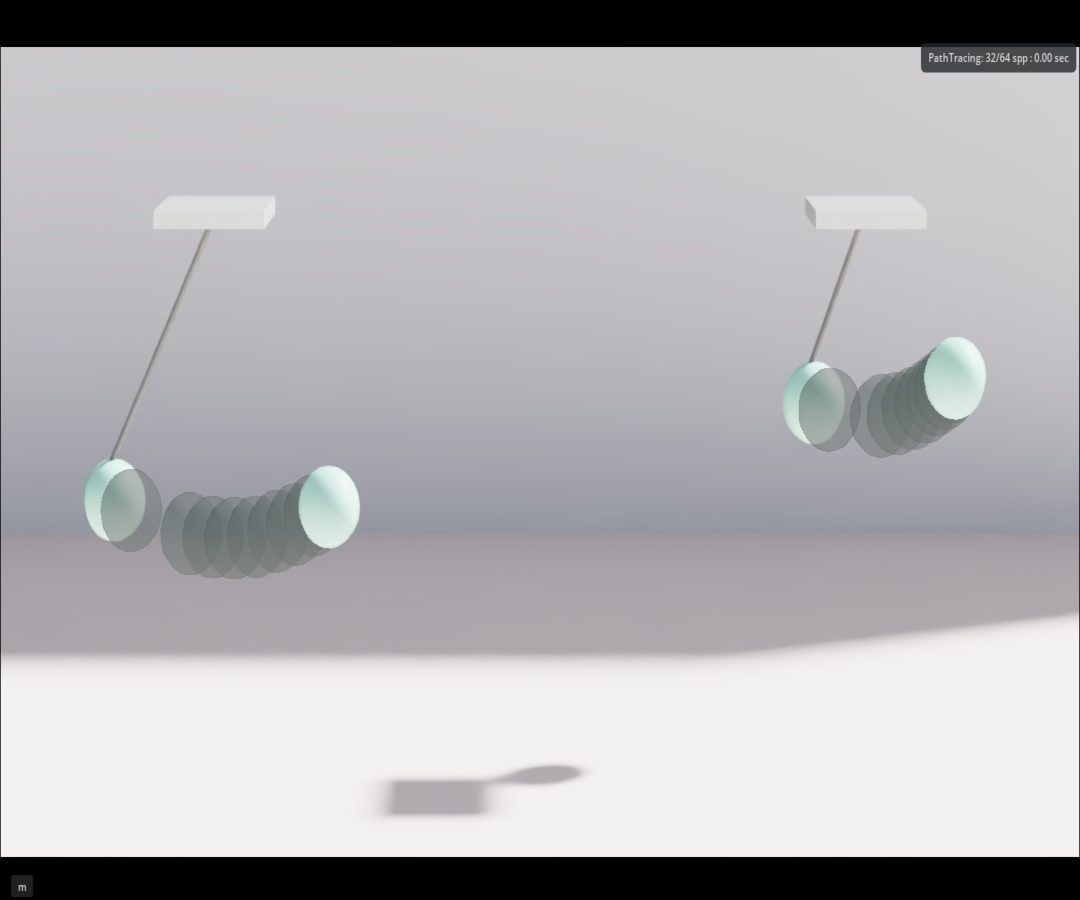}
\end{minipage}

\begin{minipage}[t]{0.75\linewidth}
\paragraph{Mass-Spring.}
\vspace{-30pt}
Two springs with identical spring constants are loaded with masses $m_1$ and $m_2$ and allowed to settle at equilibrium. By Hooke's law, equilibrium extension is $x = mg/k$, so extension scales linearly with suspended mass. The relational invariant is $x_1/x_2 = m_1/m_2$, with proportional extensions.
\end{minipage}\hfill
\begin{minipage}[c]{0.21\linewidth}
\centering
\includegraphics[
    width=\linewidth,
    trim=180px 100px 180px 80px,
    clip
]{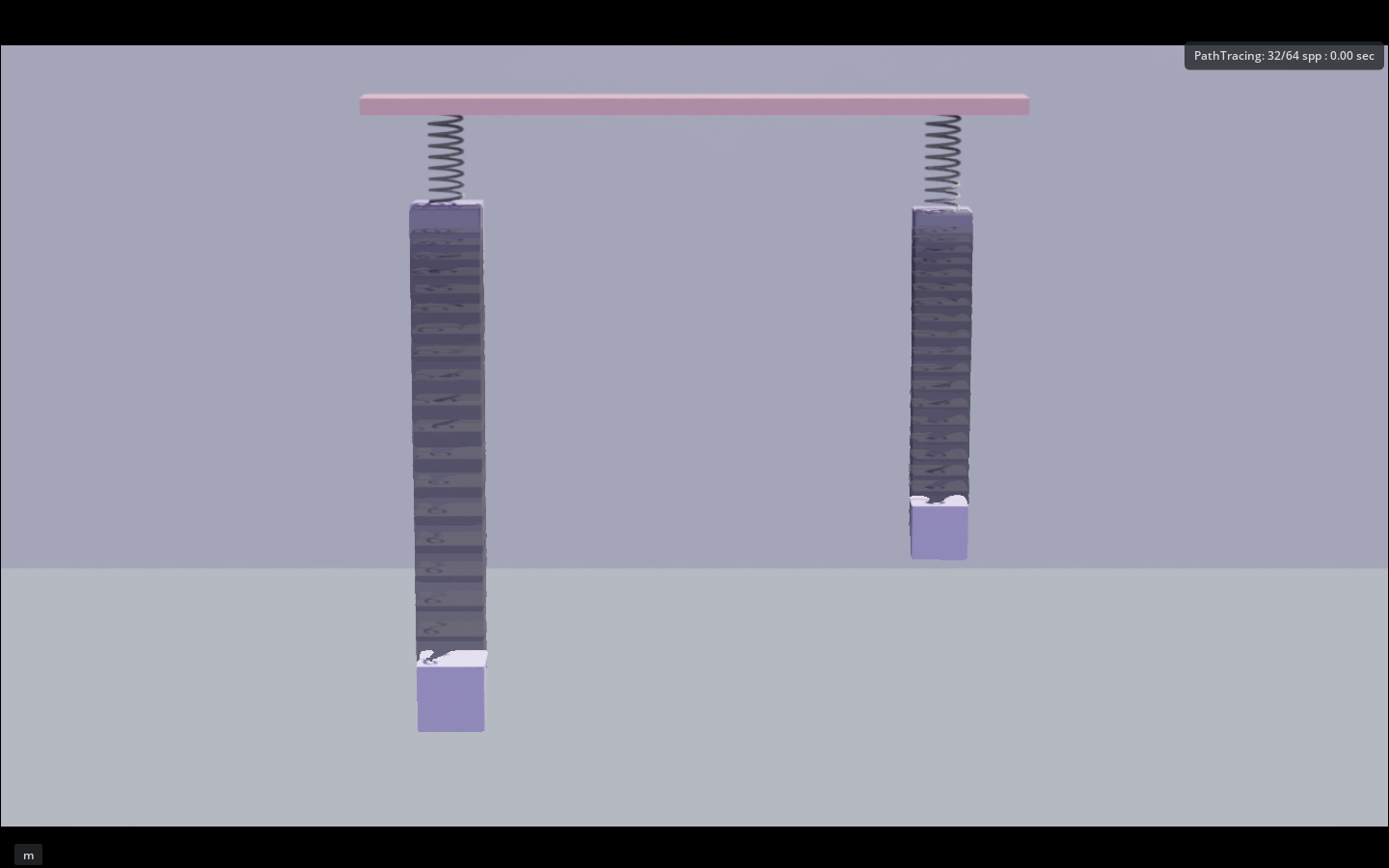}
\end{minipage}

\begin{table}[t]
\centering
\small
\caption{\textbf{Relational invariants across the eight Principia phenomena.} For each scenario, the listed constants are held fixed across paired objects and the listed variables differ. The relational invariant in the rightmost column of the form $\mathcal{F}_\phi(o_1) = \mathcal{F}_\phi(o_2)$ must hold whenever both objects obey the same physical law, regardless of camera, scale, or frame rate.}
\setlength{\tabcolsep}{4pt}
\renewcommand{\arraystretch}{1.25}
\begin{tabular}{lccc}
\toprule
\textbf{Scenario} & \textbf{Constants} & \textbf{Variables} & \textbf{Relational Invariant} \\
\midrule

Restitution
& Material, surface
& Drop heights $h_1, h_2$
& $\sqrt{r_1/h_1} = \sqrt{r_2/h_2}$ \\

Gravity
& Drop conditions
& Drop heights $h_1, h_2$
& $h_1/h_2 = (t_1/t_2)^2$ \\

Friction
& Incline angle $\theta$, friction coeff. $\mu$
& Block masses $m_1, m_2$
& $t_1 = t_2$ \\

Rotational Inertia
& Mass $m$, radius $R$, angle $\theta$
& Moment of inertia $I_1, I_2$
& $\dfrac{t_1}{t_2} = \sqrt{\dfrac{1 + I_1/mR^2}{1 + I_2/mR^2}}$ \\

Projectile
& Launch angle $\theta$, gravity $g$
& Launch heights $h_1, h_2$
& $R \propto v \propto \sqrt{h}$ \\

Momentum (height)
& Block mass $m$, ball mass
& Release heights $h_1, h_2$
& $x \propto v \propto h$ \\

Momentum (mass)
& Release height $h$, ball mass
& Block masses $m_1, m_2$
& $x \propto \frac{1}{m}$ \\

Pendulum
& Release angle $\theta$, gravity $g$
& Lengths $l_1, l_2$
& $T_1/T_2 = \sqrt{(l_1/l_2)}$ \\

Mass-Spring
& Spring constant $k$, gravity $g$
& Masses $m_1, m_2$
& $x_1/x_2 = (m_1/m_2)$ \\

\bottomrule
\end{tabular}
\label{tab:rel_ratio}
\end{table}

\section{Experiments}
\label{sec:experiments}
\subsection{Setup}
We evaluate six video generators (Omni~\cite{google_gemini_omni_flash_2026}, Veo-3.1~\cite{Google_Veo3_2025}, Wan2.2-5B/14B~\cite{wan2025}, Cosmos-2.5-2B/14B~\cite{nvidia_cosmos_predict2}) and four vision-language models (Gemini-3.1-Pro, Gemini-3-Flash~\cite{Google_Gemini_Pro_2026}, Qwen-32B, Qwen-4B~\cite{qwen3.5}). Each generator is conditioned on a text prompt and the first frame of a real recording, with the experimenter, suspension strings, and visible release mechanisms inpainted out using Nano Banana 2~\cite{google2026gemini31flashimage} so the model conditions on physics rather than apparatus. We sample each scenario using multiple random seeds and average the resulting scores across seeds to obtain a sample score.
The Principia benchmark assumes that models can generate basic qualitatively correct motion (e.g., a dropped object moves downward or an object on an inclined plane moves down the slope). We therefore filter out non-conforming videos before computing the Principia scores(More details in Appendix \ref{app:dc_filter}).
Total inference compute exceeds 2{,}600 A100-hours across the four open-weights generators (Appendix~\ref{app:compute}).

\paragraph{Principia Consistency Score.}
For each phenomenon $\phi$, the invariant defines two scalar quantities $\mathcal{F}_\phi(o_1)$ and $\mathcal{F}_\phi(o_2)$ that should be equal under correct physics. Depending on the phenomenon, these quantities may correspond either to directly measured values (friction) or to ratios derived from the measured values (gravity, spring, restitution, rotational inertia, pendulum). The specific quantities used for each phenomenon are listed in Table \ref{tab:rel_ratio}. We measure how closely a generated video satisfies the invariant using a normalized consistency score:

\vspace{-10pt}
\[
S_\phi \;=\; 1 - \frac{|\mathcal{F}_\phi(o_1) - \mathcal{F}_\phi(o_2)|}{|\mathcal{F}_\phi(o_1)| + |\mathcal{F}_\phi(o_2)|}.
\]
$S_\phi$ equals $1$ when the invariant holds exactly and decreases toward $0$ as the violation grows; concretely, $S_\phi = 0.95$ corresponds to a roughly $10\%$ relational asymmetry.
The projectile and momentum invariants is qualitative -- distance ordering rather than equality -- and is scored separately as the fraction of scenes where the ordering is satisfied. The normalization makes $S_\phi$ unit-free and bounded in $[0, 1]$; per-phenomenon scores reported throughout this paper are calculated by taking mean and standard deviation across sample scores.

\subsection{Visual Quality and Physical Fidelity Are Decoupled}
\label{sec:headline}

\begin{figure*}[t!]
  \centering
  \includegraphics[width=\textwidth,keepaspectratio]{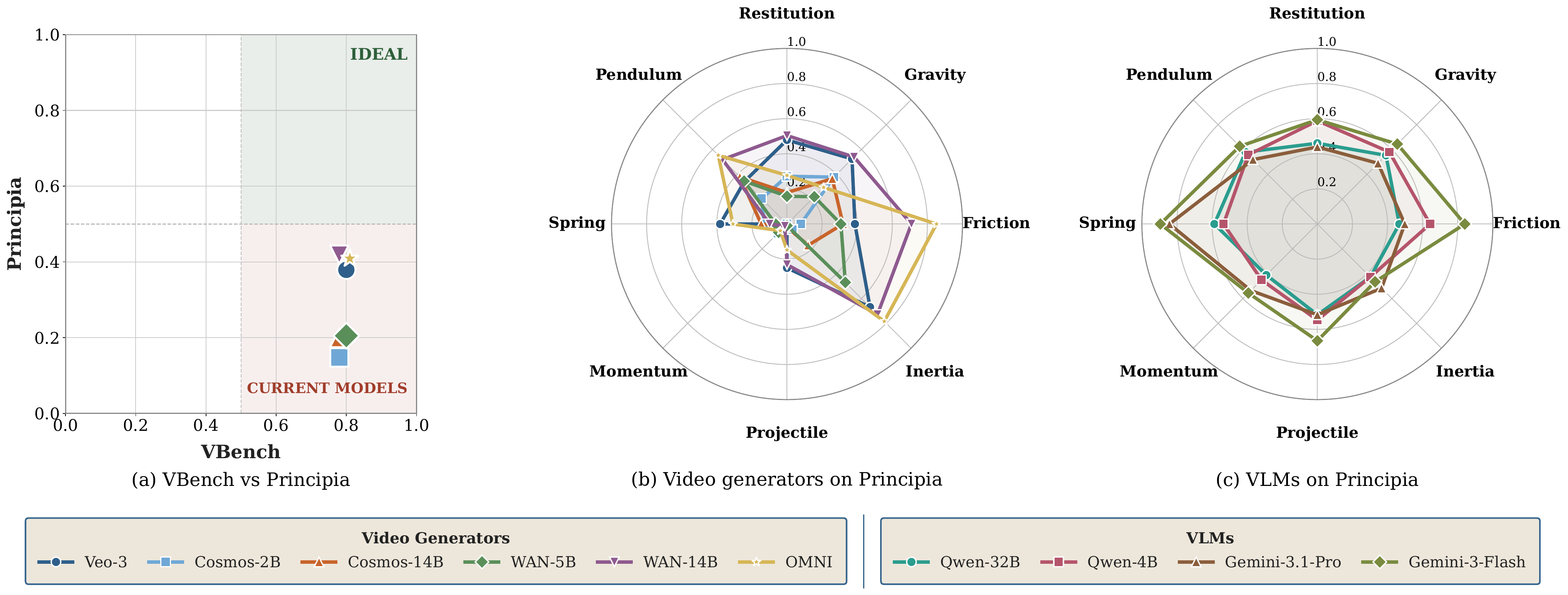}

  \caption{\textbf{Visual quality decouples from physical fidelity, and different model classes fail on different phenomena.}
  \textbf{(a)} All six video generators score around $0.8$ on VBench but below $0.5$ on Principia, clustering in the high-visual-quality, low-physics-fidelity region.
  \textbf{(b)} Per-phenomenon continuous consistency scores $S_\phi$ for video generators.
  \textbf{(c)} Per-phenomenon agreement scores for vision-language models. VLM polygons are smoother and more uniform than generator polygons (compare b vs c).}
  \label{fig:results_main}
\end{figure*}

\begin{figure*}[t!]
  \centering
  \vspace{-10pt}
\includegraphics[width=.98\textwidth,keepaspectratio]{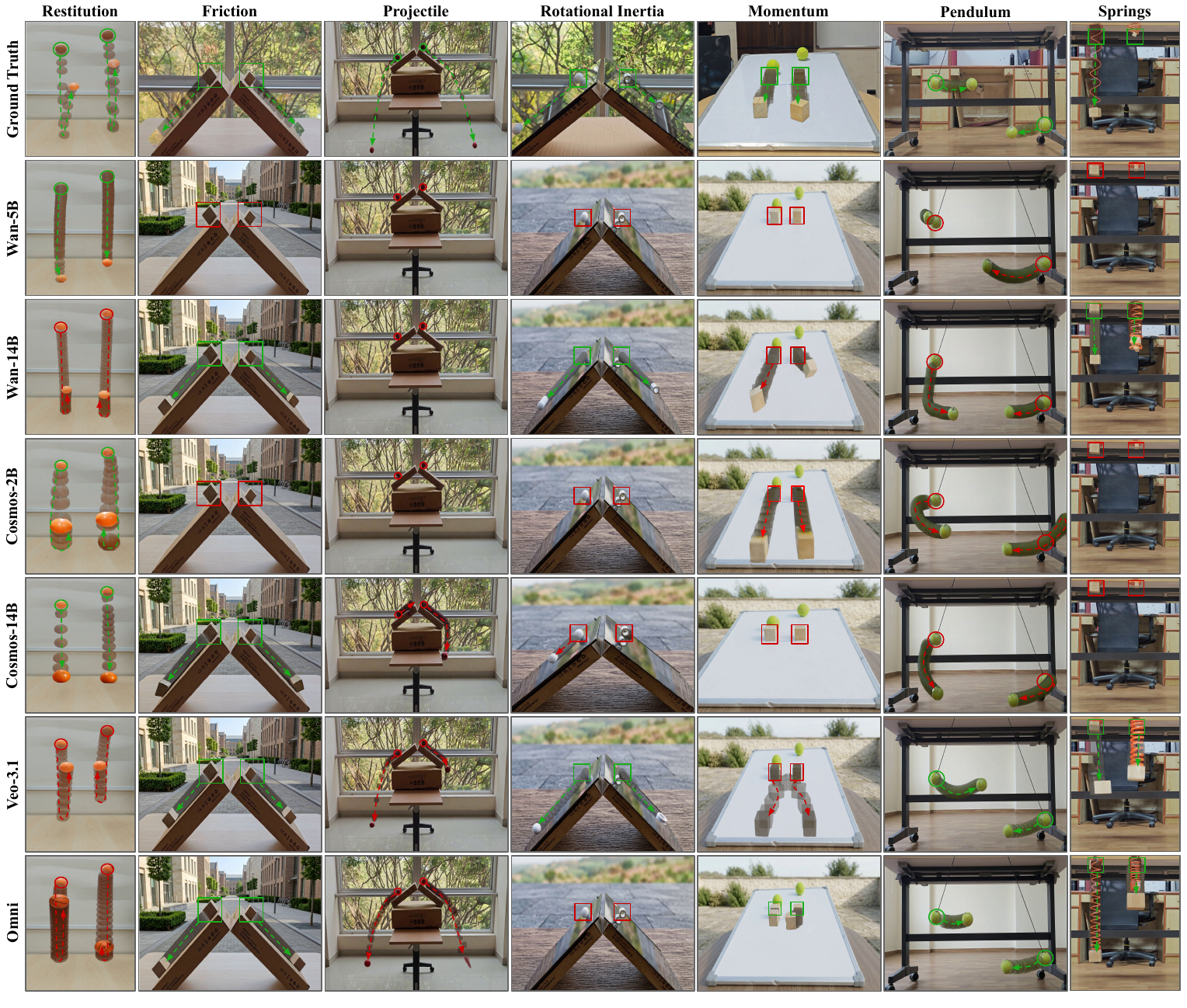}
  \vspace{-10pt}
  \caption{\textbf{Qualitative results on Principia.} 
  Stroboscopic composites of generated videos across all seven phenomena (columns) and the five evaluated video generators alongside ground truth (rows). For each phenomenon, we show two sets with different physical configurations and visual contexts. \textbf{Green markings} indicate motion that satisfies the relational invariant; \textbf{red markings} indicate physics violations. Across both sets, we include samples where the real recorded videos are edited to vary backgrounds, lighting, and scene appearance (e.g., indoor vs.\ outdoor environments, different surface textures, and illumination conditions), allowing the dataset to be scaled and diversified without requiring new physical recordings. 
  For these samples, we additionally provide the original ground-truth first frame prior to augmentation.
  }
  \label{fig:qualitative-1}
  \vspace{-20pt}
\end{figure*}

\begin{figure*}[t!]
  \centering
  \vspace{-10pt}
\includegraphics[width=.98\textwidth,keepaspectratio]{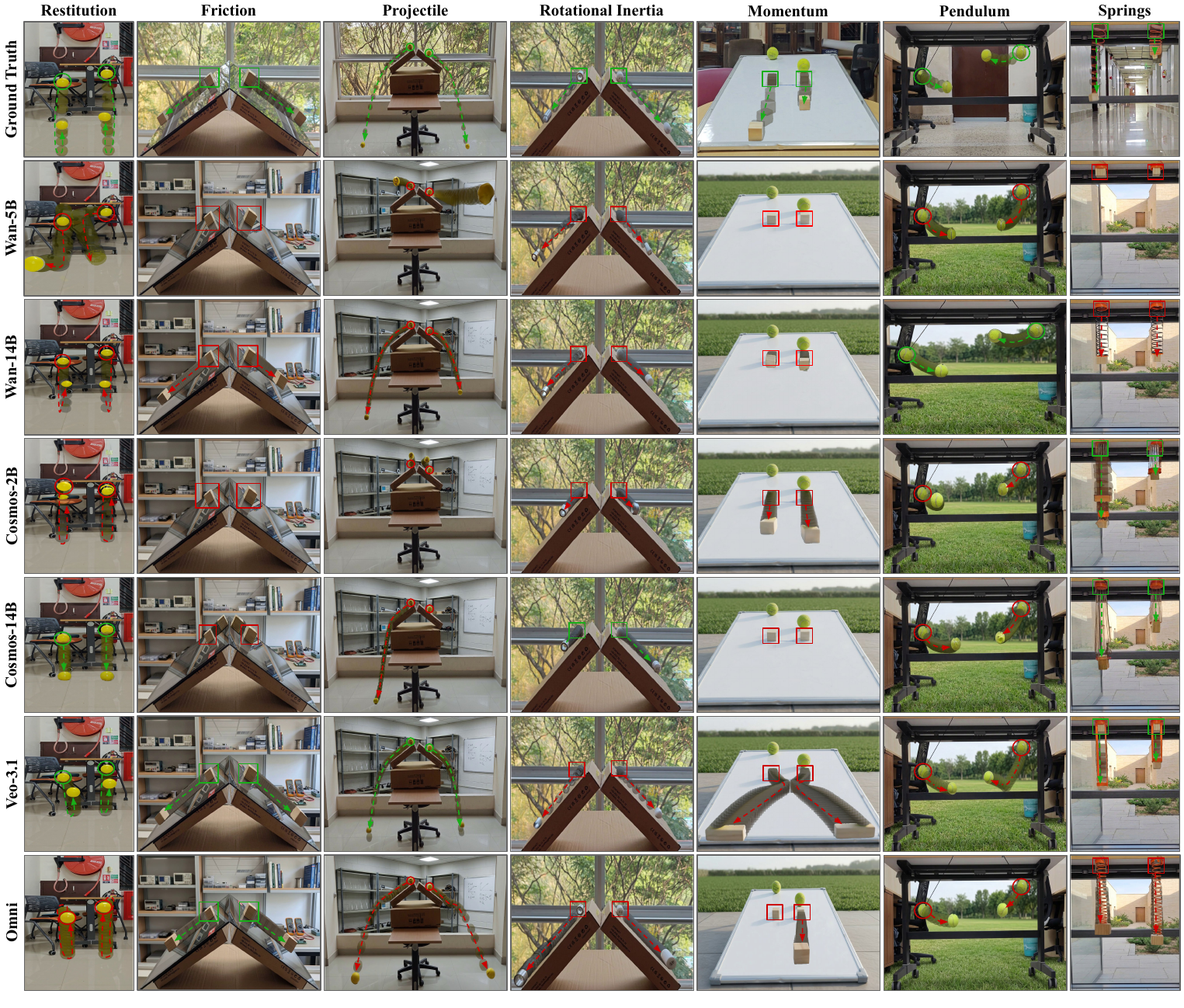}
  \vspace{-10pt}
  \caption{\textbf{Additional Qualitative results on Principia.} 
  Stroboscopic composites of generated videos across all seven phenomena (columns) and the five evaluated video generators alongside ground truth (rows). For each phenomenon, we show two sets with different physical configurations and visual contexts. \textbf{Green markings} indicate motion that satisfies the relational invariant; \textbf{red markings} indicate physics violations. Across both sets, we include samples where the real recorded videos are edited to vary backgrounds, lighting, and scene appearance (e.g., indoor vs.\ outdoor environments, different surface textures, and illumination conditions), allowing the dataset to be scaled and diversified without requiring new physical recordings. 
  For these samples, we additionally provide the original ground-truth first frame prior to augmentation.
  }
  \label{fig:qualitative-2}
  \vspace{-10pt}
\end{figure*}

Figure~\ref{fig:results_main}(a) plots VBench\cite{huang2023vbench} against Principia. All six generators cluster around $0.8$ on VBench but between $0.14$ and $0.42$ on Principia -- visual quality and physical fidelity are nearly orthogonal. Models at the visual-quality frontier are no more likely to satisfy physical invariants than substantially lower-quality alternatives.

\subsection{Per-Phenomenon Results: Generators}

\begin{table*}[t!]
\centering
\small
\caption{\textbf{Performance of video generators on Principia.} We report the continuous relational consistency score $S_\phi$ for each physical phenomenon as mean $\pm$ standard deviation across samples, where each samples score is averaged across multiple seeds. Higher values indicate better adherence to the underlying physical invariant. \textbf{Bold} indicates the best-performing model for each phenomenon. The Overall score reports the mean $\pm$ standard deviation across scenario scores and summarizes the model's general performance across physical phenomena.}
\label{tab:results_real}
\setlength{\tabcolsep}{4pt}
\renewcommand{\arraystretch}{1.0}
\resizebox{\textwidth}{!}{%
\begin{tabular}{lccccccccc}
\toprule
\textbf{Model} & \textbf{Restitution} & \textbf{Gravity} & \textbf{Friction} & \textbf{Inertia} & \textbf{Projectile} & \textbf{Momentum} & \textbf{Spring} & \textbf{Pendulum} & \textbf{Overall} \\
\midrule

Veo-3.1~\cite{Google_Veo3_2025}
& 0.479 $\pm$ 0.354
& 0.524 $\pm$ 0.378
& 0.387 $\pm$ 0.400
& 0.669 $\pm$ 0.380
& \textbf{0.250 $\pm$ 0.370}
& 0.000 $\pm$ 0.000
& \textbf{0.381 $\pm$ 0.160}
& 0.343 $\pm$ 0.309
& 0.379 $\pm$ 0.186 \\

\midrule

Cosmos-2.5-2B~\cite{nvidia_cosmos_predict2}
& 0.272 $\pm$ 0.301
& 0.378 $\pm$ 0.357
& 0.078 $\pm$ 0.176
& 0.044 $\pm$ 0.154
& 0.000 $\pm$ 0.000
& \textbf{0.074 $\pm$ 0.177}
& 0.130 $\pm$ 0.167
& 0.205 $\pm$ 0.191
& 0.148 $\pm$ 0.120 \\

\midrule

Cosmos-2.5-14B~\cite{nvidia_cosmos_predict2}
& 0.179 $\pm$ 0.233
& 0.365 $\pm$ 0.337
& 0.319 $\pm$ 0.339
& 0.171 $\pm$ 0.303
& 0.019 $\pm$ 0.094
& 0.015 $\pm$ 0.084
& 0.147 $\pm$ 0.161
& 0.380 $\pm$ 0.268
& 0.199 $\pm$ 0.135 \\

\midrule

Wan2.2-5B~\cite{wan2025}
& 0.158 $\pm$ 0.243
& 0.220 $\pm$ 0.262
& 0.307 $\pm$ 0.357
& 0.470 $\pm$ 0.371
& 0.009 $\pm$ 0.067
& 0.067 $\pm$ 0.249
& 0.065 $\pm$ 0.125
& 0.346 $\pm$ 0.244
& 0.205 $\pm$ 0.150 \\

\midrule

Wan2.2-14B~\cite{wan2025}
& \textbf{0.504 $\pm$ 0.338}
& \textbf{0.540 $\pm$ 0.344}
& 0.709 $\pm$ 0.291
& 0.729 $\pm$ 0.260
& 0.231 $\pm$ 0.300
& 0.017 $\pm$ 0.090
& 0.099 $\pm$ 0.150
& 0.519 $\pm$ 0.235
& \textbf{0.419 $\pm$ 0.253} \\

\midrule

Omni~\cite{google_gemini_omni_flash_2026}
& 0.277 $\pm$ 0.327
& 0.295 $\pm$ 0.334
& \textbf{0.852 $\pm$ 0.264}
& \textbf{0.784 $\pm$ 0.261}
& 0.148 $\pm$ 0.266
& 0.054 $\pm$ 0.155
& 0.308 $\pm$ 0.205
& \textbf{0.554 $\pm$ 0.241}
& 0.409 $\pm$ 0.272 \\

\bottomrule
\end{tabular}%
}
\vspace{-8pt}
\end{table*}

Table~\ref{tab:results_real} reports per-phenomenon results. \textit{No generator exceeds an average consistency score of $0.5$}, highlighting the difficulty of maintaining relational physical consistency across diverse scenarios. Wan2.2-14B achieves the best overall performance with an average score of $0.419$, narrowly outperforming Omni and Veo-3.1 at $0.409$ and $0.38$ respectively despite their closed-frontier status.
\textit{Within-family scaling trends are highly uneven}. Scaling Cosmos-2.5 from 2B to 14B produces modest gains in friction ($+0.24$), inertia ($+0.13$), and pendulum consistency ($+0.18$), while reducing performance on restitution ($-0.09$), gravity ($-0.01$), and momentum ($-0.06$). Similarly, scaling Wan2.2 from 5B to 14B produces substantial gains across restitution ($+0.35$), gravity ($+0.32$), friction ($+0.40$), inertia ($+0.26$), and projectile ($+0.22$), with smaller gains on spring ($+0.03$) and pendulum ($+0.17$), while momentum consistency slightly decreases ($-0.05$). These results suggest that larger model scale does not uniformly improve physical reasoning, and that different invariants stress distinct failure modes in current video generators.
Compute also fails to predict fidelity -- Cosmos-2.5-14B uses more compute per video than Wan2.2-14B (74 vs 66 min) yet scores $0.22$ lower (Appendix~\ref{app:model_config}).

Figure~\ref{fig:results_main}(b) visualizes per-model failure profiles. Each polygon exhibits a distinct performance profile rather than uniform weakness. Omni performs particularly well on inertia and friction but struggles with restitution, gravity, and momentum. Wan2.2-14B achieves the strongest overall performance, reaching the performance frontier among evaluated models across all scenarios except spring and momentum. The Cosmos models exhibit smaller polygon areas, indicating weaker physical consistency across phenomena. All models perform poorly on the momentum task, potentially reflecting the greater difficulty of modeling complex multi-object interactions. Notably, no model dominates across all physical phenomena, with none fully covering the radar chart. Qualitative comparisons appear in Figures~\ref{fig:qualitative-1},\ref{fig:qualitative-2}.

\subsection{Vision-Language Models}
\label{sec:vlms}

\begin{table}[t]
\centering
\small
\caption{\textbf{Performance of vision-language models} Each cell reports the agreement score $A_\phi$: the fraction of scenes where the VLM's PASS/FAIL judgment matches the binary ground truth. Higher is better. \textbf{Bold} denotes the best per column. The Overall column reports the mean across scenarios for each model.}
\label{tab:results_vlms}
\setlength{\tabcolsep}{4pt}
\resizebox{\columnwidth}{!}{%
\begin{tabular}{lccccccccc}
\toprule
\textbf{Model} &
\textbf{Restitution} &
\textbf{Gravity} &
\textbf{Friction} &
\textbf{Inertia} &
\textbf{Projectile} &
\textbf{Momentum} &
\textbf{Spring} &
\textbf{Pendulum} &
\textbf{Overall} \\
\midrule

Gemini-3.1-Pro~\cite{Google_Gemini_Pro_2026} &
0.439 &
0.489 &
0.498 &
\textbf{0.517} &
0.517 &
0.536 &
0.843 &
0.519 &
0.545 \\

Gemini-3-Flash~\cite{Google_Gemini_Pro_2026} &
\textbf{0.594} &
\textbf{0.644} &
\textbf{0.840} &
0.465 &
\textbf{0.665} &
\textbf{0.556} &
\textbf{0.895} &
\textbf{0.626} &
\textbf{0.661} \\

Qwen-32B~\cite{qwen3.5} &
0.460 &
0.553 &
0.467 &
0.427 &
0.517 &
0.411 &
0.587 &
0.579 &
0.500 \\

Qwen-4B~\cite{qwen3.5} &
0.588 &
0.580 &
0.642 &
0.427 &
0.547 &
0.450 &
0.535 &
0.557 &
0.541 \\

\bottomrule
\end{tabular}%
}
\vspace{-8pt}
\end{table}

VLMs are evaluated on whether they can \emph{detect} relational physics violations rather than \emph{generate} physically correct motion. We evaluate the models on a dataset comprising both real-world samples and synthetic samples from the Principia-Synth dataset. Principia-Synth consists of rendered videos depicting both real-world physics and physics violations for each phenomenon. The anti-physics videos are constructed by explicitly violating the relational invariant associated with each phenomenon. For every scene, the VLM receives the full video along with a phenomenon-specific PASS/FAIL prompt (Appendix~\ref{app:vlm_protocol}); its response is then compared against the known ground truth. Table~\ref{tab:results_vlms} reports the agreement scores for each phenomenon. No VLM achieves an average agreement exceeding $0.7$, indicating \emph{near-chance performance} and suggesting architectural, rather than scale-related, limitations.

\section{Discussion and Conclusion}

Video generators are increasingly marketed as world models. The results on \emph{Principia} suggest otherwise: across six state-of-the-art generators, no model exceeds $0.45$ on the continuous metric despite all scoring around $0.8$ on visual quality benchmarks; scaling within an architecture regresses on at least one phenomenon in both the Cosmos and Wan families; and open-weights exceeds closed-frontier (Wan2.2-14B exceeds Veo-3.1 and Omni). Current generators produce highly realistic textures and visually appealing renderings but fail to enforce the relational invariants that physical laws require. They have learned to render the world without learning what holds it together.

Vision-language models are no better at \emph{detecting} physical consistency than video generators are at producing it. Most models perform near chance level, with even the best-performing model achieving only 67\% accuracy in distinguishing physically consistent videos from those containing physics violations.

\paragraph{Scope and limitations.} \emph{Principia} tests macroscopic Newtonian mechanics; it does not address fluid dynamics, soft-body deformation, or thermodynamic phenomena. Generators sometimes produce videos with hallucinated objects, missing objects, or severe deformations -- failures unrelated to physical reasoning. We do not automatically separate these from genuine physics violations; our scope is physical-law adherence in videos that are otherwise interpretable. As generators mature, more of their output will be evaluable under Principia without changes to the benchmark itself.

\paragraph{Resources and outlook.} Final-scale evaluation generated over $4{,}000$ videos at approximately $2{,}600$ A100-hours, with comparable additional compute consumed during prompt iteration, threshold calibration, and apparatus development. Because the consistency score is computed within each generated video, the benchmark is robust to test-set contamination. The first frames are precisely staged physical setups that current image generators cannot reproduce, but they can be edited cheaply to vary backgrounds and lighting using modern image editing models~\cite{google2026gemini31flashimage}, enabling extension to new visual setups. We  release first-frame inputs and scenario prompts; full real-world videos will be held back to preserve benchmark integrity. If video generators are to serve as world simulators, they must preserve the structural constraints that physical laws impose---not merely the appearance of motion. Closing this gap will likely require new training signals or architectural biases for relational invariants.

{\small
\bibliographystyle{plain}
\bibliography{main}
}

\clearpage
\section*{Appendix}
\appendix

\section{Video Generator Evaluation}
Video generators are evaluated on their ability to \emph{generate} videos satisfying the relational constraint of the phenomenon. In this section we provide additional details on the Principia benchmark and evaluation protocol.

\subsection{Dataset Statistics:}
\begin{figure*}[ht]
    \centering
    \includegraphics[
        width=\textwidth,
        keepaspectratio
    ]{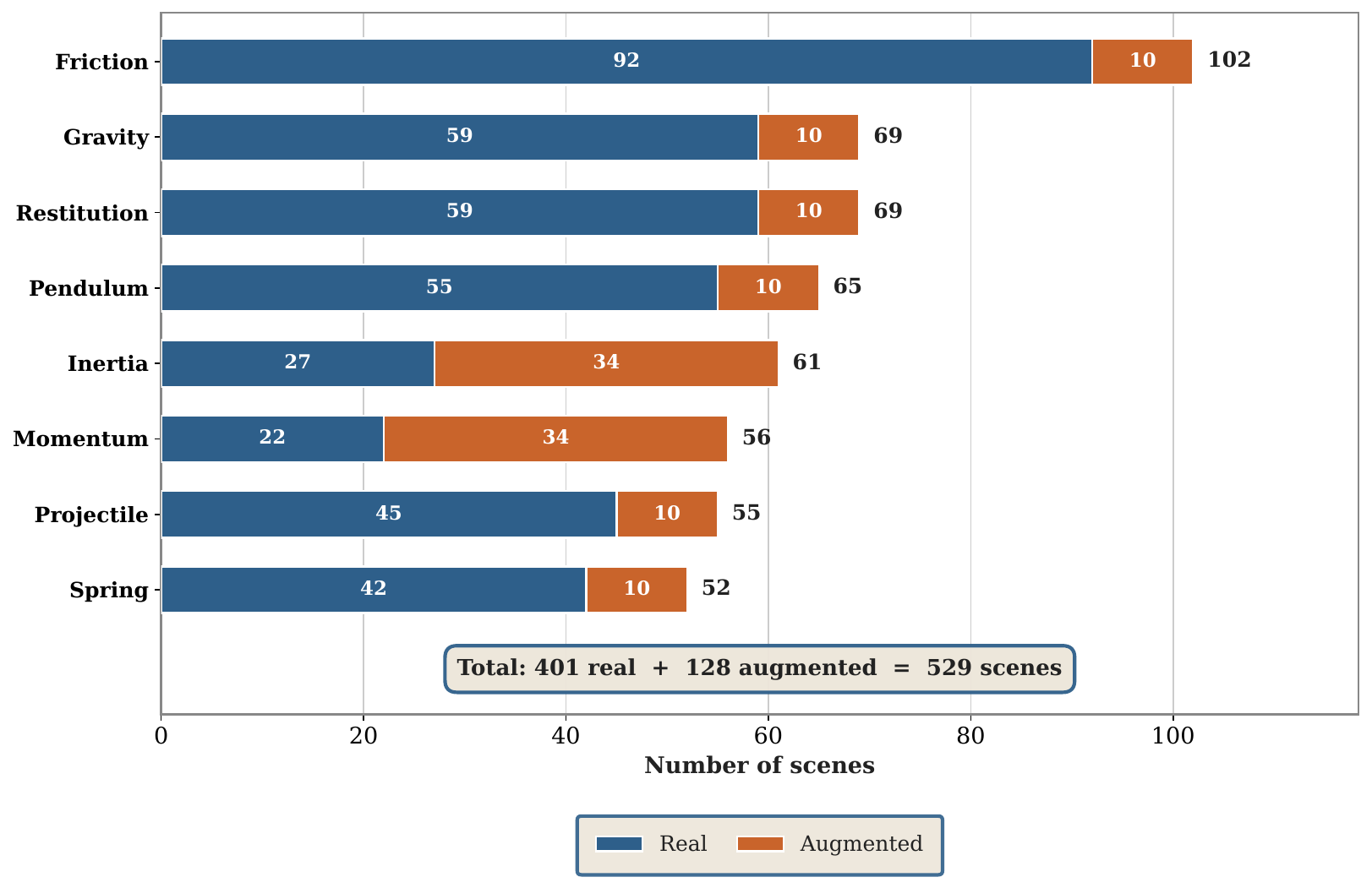}
    \vspace{-8pt}
    \caption{\textbf{Dataset statistics across physical phenomena.}
    The figure summarizes the number of samples available for each physical phenomenon, separated into real-world and augmented scenarios.}
    \label{fig:dataset_statistics}
\end{figure*}

The dataset consists of 401 real-world scenes captured in-house. For each scene, the first frame is extracted and used along with a text prompt as the conditioning for video generation. The conditioning frames are first edited using image-editing models~\cite{google2026gemini31flashimage} to remove the human experimenter and experimental apparatus. These edited frames are then further augmented to increase visual diversity, resulting in a total of 529 scenes. The per-scenario composition of the dataset is shown in Fig.~\ref{fig:dataset_statistics}.

\subsection{Evaluation protocol}

All quantitative measurements are extracted automatically from the generated videos. Object trajectories are obtained using SAM3\cite{carion2025sam3}, which tracks the position(centroid/bottom) of each object throughout the sequence. Phenomenon-specific events (e.g., impact, arrival, turning points) are then detected from the tracked trajectories to compute the quantities required for evaluating the corresponding relational invariant.

\paragraph{Restitution and Gravity.}
The centroid trajectory of each ball is tracked using SAM3. Ground impact is detected when the centroid falls below a predefined ground threshold and the vertical velocity changes sign from positive (downward) to negative (upward). The rebound apex is identified as the subsequent frame where the vertical velocity changes from negative to positive. Initial drop height, rebound height, and time of flight are computed from these detected events and used to evaluate the restitution and gravity invariants.

\paragraph{Friction and Rotational Inertia.}
For both scenarios, the arrival time of each object is measured as the first frame in which its bottom most pixel(to handle size differences in blocks) crosses a predefined annotated threshold at the end of the incline. These arrival times are used to evaluate the friction and rotational invariants.

\paragraph{Projectile Motion.}
The projectile centroid trajectory is tracked using SAM3. Ground impact is detected when the vertical velocity becomes positive after descent and the centroid crosses the ground threshold. The horizontal range is computed as the horizontal distance between the impact location and the annotated launch point at the end of the incline.

\paragraph{Pendulum.}
The oscillation period of each pendulum is estimated from the centroid trajectory of the bob. A half-period is detected when the horizontal velocity reverses sign after the bob has crossed the pendulum's vertical midpoint (anchor line). The full period is obtained by doubling the measured half-period.

\paragraph{Spring.}
The maximum extension of each spring is measured from the first oscillation. The oscillation turning point is detected when the vertical velocity changes from positive to negative, and the corresponding vertical displacement is recorded. Object masses are estimated from the cube side lengths assuming constant density, i.e., $m \propto l^3$, and the measured displacement ratio is compared against the expected mass ratio.

\paragraph{Momentum.}
The vertical displacement of each target block is measured by tracking the bottom edge of the block. The maximum post-collision displacement is extracted from the trajectory and compared across the two interactions to evaluate the expected ordering of momentum transfer.

\begin{table}[H]
\centering
\scriptsize
\caption{\textbf{Model details.} We use default configurations for high-quality generation. Inference time is measured on an A100 80GB; Veo-3.1 and Omni are accessed through their public APIs.}
\label{tab:model_Details}
\setlength{\tabcolsep}{3pt}
\renewcommand{\arraystretch}{0.9}
\resizebox{\columnwidth}{!}{%
\begin{tabular}{lcccc}
\toprule
\textbf{Model} & \textbf{Resolution} & \textbf{Frames} & \textbf{FPS} & \textbf{Time} \\
\midrule
Wan2.2-5B~\cite{wan2025} 
& 1280$\times$704 & 121 & 24 & 7 min \\

Wan2.2-14B~\cite{wan2025} 
& 1280$\times$720 & 81 & 16 & 66 min \\

Cosmos-2.5-2B~\cite{nvidia_cosmos_predict2} 
& 1280$\times$704 & 93 & 16 & 15 min \\

Cosmos-2.5-14B~\cite{nvidia_cosmos_predict2} 
& 1280$\times$704 & 93 & 16 & 74 min \\

Veo-3.1~\cite{Google_Veo3_2025} 
& 1280$\times$720 & 96 & 24 & API \\

Omni~\cite{google_gemini_omni_flash_2026}
& 1280$\times$720 & 240 & 24 & API \\
\bottomrule
\end{tabular}%
}
\vspace{-8pt}
\end{table}

\subsection{Model Configurations.}
\label{app:model_config}
We use the default configuration of each evaluated model to obtain high-quality generations. Resolution, frame count, and frame rate are reported in Table~\ref{tab:model_Details}. Inference times are measured on an NVIDIA A100 80GB and refer to wall-clock time per generated video. Veo-3.1 and Omni are accessed through public APIs.

\subsection{Prompts Structure}

We list the prompts used to generate evaluation videos with Principia. Each video generation model is conditioned on the first frame of the video along with the corresponding prompt. 

\begin{itemize}[leftmargin=*]

\item \textbf{Restitution/Gravity:} \textit{A video showing two identical balls being dropped from two different heights onto the ground. The camera is static and positioned to clearly capture the vertical motion of both balls. Both balls fall naturally under gravity, accelerating freely with no air resistance and hit the ground. The balls bounce a few times on the ground before coming to rest.}

\item \textbf{Friction:} \textit{A video showing two identical inclined planes with the same angle of inclination. Two blocks of the same material are placed on the inclines at the same height from the bottom. The blocks start from rest and slide down the planes. The camera is fixed and motion is strictly along the planes.}

\item \textbf{Rotational Inertia:} \textit{A video showing two identical inclined planes with the same angle of inclination. A solid cylinder is placed on one plane and a hollow cylinder of the same mass and outer radius is placed on the other. Both cylinders start from rest at the same height and roll down their planes without slipping. The camera is fixed and the motion is strictly along the planes.}

\item \textbf{Projectile:} \textit{A video showing two identical inclined planes with the same angle of inclination positioned on an elevated surface. Two identical balls are placed on the inclines. The balls start from rest, roll down the planes, and launch off the edge to the ground. The camera is fixed and captures the complete motion of the balls.}

\item \textbf{Momentum:} \textit{A video from a fixed camera showing a single wide inclined plane facing directly towards the camera. Two identical tennis balls are placed on the incline at different/same starting heights. Both balls start from rest and roll down the plane towards the camera without slipping. Two solid wooden square blocks of same/different masses are placed at the same height from the bottom, directly in the path of each ball. The balls collide head-on with their respective wooden blocks, causing them to move forward. The motion is strictly along the plane and the horizontal direction after the collision.}

\item \textbf{Mass-Spring:} \textit{A video from a fixed camera showing two vertical spring-mass systems suspended side-by-side from a fixed horizontal support. Both springs are entirely identical, having the same unstretched length and the same spring constant. Two wooden cube blocks having different masses are attached to the springs. Initially, both blocks are held elevated right up near the horizontal support, fully compressing the springs. At time zero, both blocks are released simultaneously from rest. The blocks fall under gravity, stretching the springs downwards, and then oscillate vertically about their respective equilibrium positions. The camera is fixed and clearly captures the continuous vertical motion of the system.}

\item \textbf{Pendulum:} \textit{A video from a fixed camera showing two simple pendulums suspended from the same horizontal support. The pendulum bobs are identical in size, shape, and material, but are attached to strings of different lengths. Both pendulums are initially displaced to the same angular position and released simultaneously from rest without any initial push. The pendulums swing freely under gravity in a plane parallel to the camera. The camera is fixed and captures the complete oscillatory motion of both pendulums.}

\end{itemize}

\subsection{Evaluation Compute.}
\label{app:compute}
Generating the corpus required substantial compute. Wan2.2-14B alone required approximately 1,100 A100-hours (46 days), while Cosmos-2.5-14B required approximately 1300 A100-hours (54 days) to generate the 529 scenes across two seeds. Across all four open-weight models, inference required approximately 2,800 A100-hours, equivalent to 120 days of continuous compute on a single A100.

\subsection{Directional Consistency Score.}
\label{app:dc_filter}
We define a Directional Consistency Score(DCS) that evaluates the generated object motion trajectory against an expected object motion trajectory to check for basic qualitative motion. 

\begin{equation}
\mathrm{DCS}
=
\frac{1}{|S|}
\sum_{i\in S}
\frac{\Delta \mathbf{p}_i^\top \hat{\mathbf{d}}}
{\|\Delta \mathbf{p}_i\|},
\end{equation}

where $\Delta \mathbf{p}_i = \mathbf{p}_{i+1} - \mathbf{p}_i$ is the object's displacement between consecutive frames, $\hat{\mathbf{d}}$ is the unit vector representing the expected direction of motion, and $S = \{\,i : \|\Delta \mathbf{p}_i\| \ge \epsilon\,\}$ is the set of frames with non-negligible displacement. Since both $\Delta \mathbf{p}_i / \|\Delta \mathbf{p}_i\|$ and $\hat{\mathbf{d}}$ are unit vectors, $\mathrm{DCS} \in [-1,1]$, where a value of $1$ indicates perfect agreement with the expected direction and $-1$ indicates motion consistently opposite to the expected direction.

In all scenarios except the pendulum, $p_i$ denotes the vertical ($y$) coordinate of the tracked object, with the expected direction $\hat{\mathbf{d}}=+1$ corresponding to downward motion in image coordinates. For the pendulum scenario, $p_i$ denotes the horizontal ($x$) coordinate, and $\hat{\mathbf{d}}$ is defined toward the equilibrium position (the center). We use a Directional Consistency Score threshold of $0.8$ to filter non-conformational videos. Videos with a DCS below this threshold are assigned a Principia Score of $0$, thereby penalizing models that fail to generate the expected qualitative object motion.

\subsection{Scaling Within Architecture}

\begin{figure}[h]
\centering
\includegraphics[width=\linewidth]{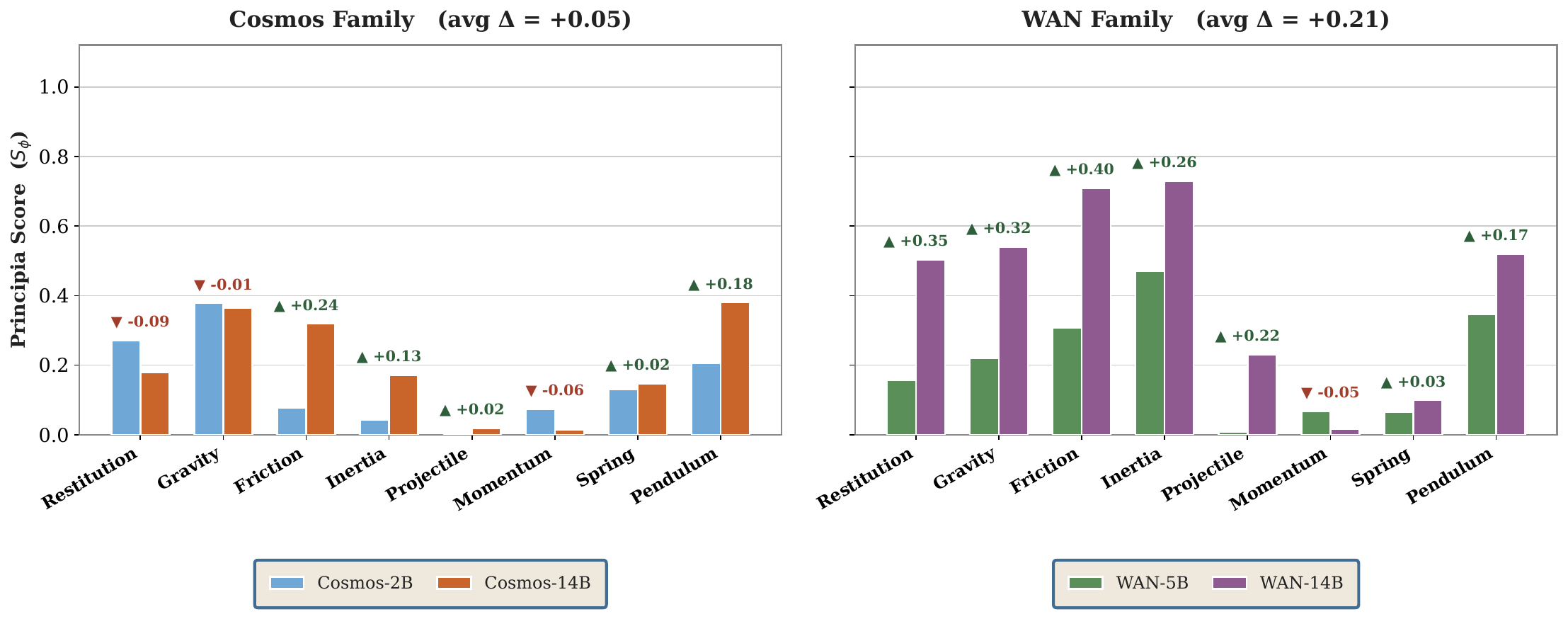}
\caption{\textbf{Within-architecture scaling effects.} 
Per-phenomenon score changes when scaling Cosmos from 2B to 14B parameters (left) and Wan2.2 from 5B to 14B (right). 
Annotations indicate per-phenomenon delta; red values denote regression. 
Both families show substantial gains on some phenomena (friction, pendulum) but regression on others (Cosmos restitution, Wan momentum), indicating that scaling does not uniformly improve physical fidelity.}
\label{fig:supp_scaling}
\end{figure}

Figure~\ref{fig:supp_scaling} compares performance within each model family as parameters increase: Cosmos-2B to Cosmos-14B ($7\times$ scale) and Wan2.2-5B to Wan2.2-14B ($2.8\times$ scale). The annotations show the per-phenomenon delta, with negative values shown in red.

The picture is more nuanced than "scaling helps." Cosmos shows substantial improvement on friction ($+0.29$) and pendulum ($+0.24$), but \emph{regresses} on restitution ($-0.09$). Wan shows large improvements on friction ($+0.40$), inertia ($+0.26$), and pendulum ($+0.17$), but \emph{regresses} on momentum ($-0.05$). Critically, in both families at least one phenomenon \emph{degrades} with scale, suggesting that scaling within a fixed architecture does not uniformly improve physical fidelity—and may actively hurt specific physical regimes that the smaller model handled adequately.

This finding has implications for the path forward: physical fidelity is unlikely to be solved by scale alone within current architectural choices.

\subsection{Camera Sensitivity Analysis}
Accurate measurement of the relational invariants using SAM masks requires videos with a static camera, as camera motion can introduce apparent object motion and interfere with mask-based measurements. In an uncontrolled scene, deviations from the expected trajectory can arise from several confounding factors, making it difficult to attribute the measured error to physics alone. We therefore explicitly instruct the video generators to use a fixed camera. However, qualitative inspection of the generated videos shows that models do not always follow this instruction. We consequently evaluate the robustness of our evaluation protocol to camera motion. Specifically, we first compute Principia scores on our filtered real-world samples with a static camera. We then apply synthetic translational camera motion to these videos to simulate camera movement and recompute the scores. The results are presented in Tab.~\ref{app:camera}. Six out of the seven phenomena remain unaffected. The one exception - Projectile is sensitive to horizontal pan, for a reason specific to how it is scored. It is our only invariant evaluated as a horizontal range ordering rather than a continuous ratio, so a horizontal pan compresses one sphere's apparent range relative to the other and flips the ordering outright, with no partial credit. The other scenarios remain unaffected indicating that the low principia scores for the models are attributable to bad physics and not to camera motion. 

\begin{table}[ht]
\centering
\small
\caption{\textbf{Robustness to camera motion.} Principia scores on real-world videos under simulated camera motion. Pan and zoom magnitudes denote the percentage change applied to the original video.}
\label{tab:camera}
\setlength{\tabcolsep}{4pt}
\renewcommand{\arraystretch}{1.1}
\begin{tabular}{lcccc}
\toprule
\textbf{Phenomenon}
& \textbf{Original}
& \textbf{Pan Down}
& \textbf{Pan Left}
& \textbf{Zoom} \\
& 
& \textbf{2\% / 5\% / 10\%}
& \textbf{2\% / 5\% / 10\%}
& \textbf{2\% / 5\% / 10\%} \\
\midrule

Gravity
& 0.934
& 0.934 / 0.936 / 0.937
& 0.934 / 0.934 / 0.934
& 0.934 / 0.934 / 0.935 \\

Restitution
& 0.977
& 0.960 / 0.960 / 0.961
& 0.976 / 0.959 / 0.977
& 0.960 / 0.961 / 0.961 \\

Pendulum
& 0.901
& 0.899 / 0.903 / 0.901
& 0.900 / 0.912 / 0.920
& 0.903 / 0.904 / 0.902 \\

Inertia
& 0.936
& 0.935 / 0.938 / 0.901
& 0.940 / 0.900 / 0.898
& 0.936 / 0.933 / 0.900 \\

Friction
& 0.963
& 0.964 / 0.964 / 0.964
& 0.964 / 0.964 / 0.964
& 0.964 / 0.964 / 0.964 \\

Momentum
& 1.000
& 1.000 / 1.000 / 1.000
& 1.000 / 1.000 / 1.000
& 0.950 / 1.000 / 0.950 \\

Spring
& 0.881
& 0.855 / 0.867 / 0.837
& 0.844 / 0.851 / 0.850
& 0.849 / 0.850 / 0.867 \\

Projectile
& 1.000
& 1.000 / 1.000 / 1.000
& 0.861 / 0.722 / 0.417
& 1.000 / 1.000 / 0.972 \\

\bottomrule
\label{app:camera}
\end{tabular}
\end{table}

\section{Vision-Language Model Evaluation}
\label{app:vlm}

We evaluated four vision-language models---Qwen-4B, Qwen-32B, Gemini-3-Flash, and Gemini-3.1-Pro---on Principia-synth. Unlike video generators, which we score on whether their generated motion preserves relational invariants, VLMs are scored on whether they correctly identify physical violations in videos shown to them.

\subsection{Principia-Synth Anti Physics Scenarios}
\begin{figure}[ht]
\centering

\begin{subfigure}[t]{0.24\linewidth}
    \centering
    \includegraphics[
    width=\linewidth,
    trim=180px 50px 180px 130px,
    clip
]{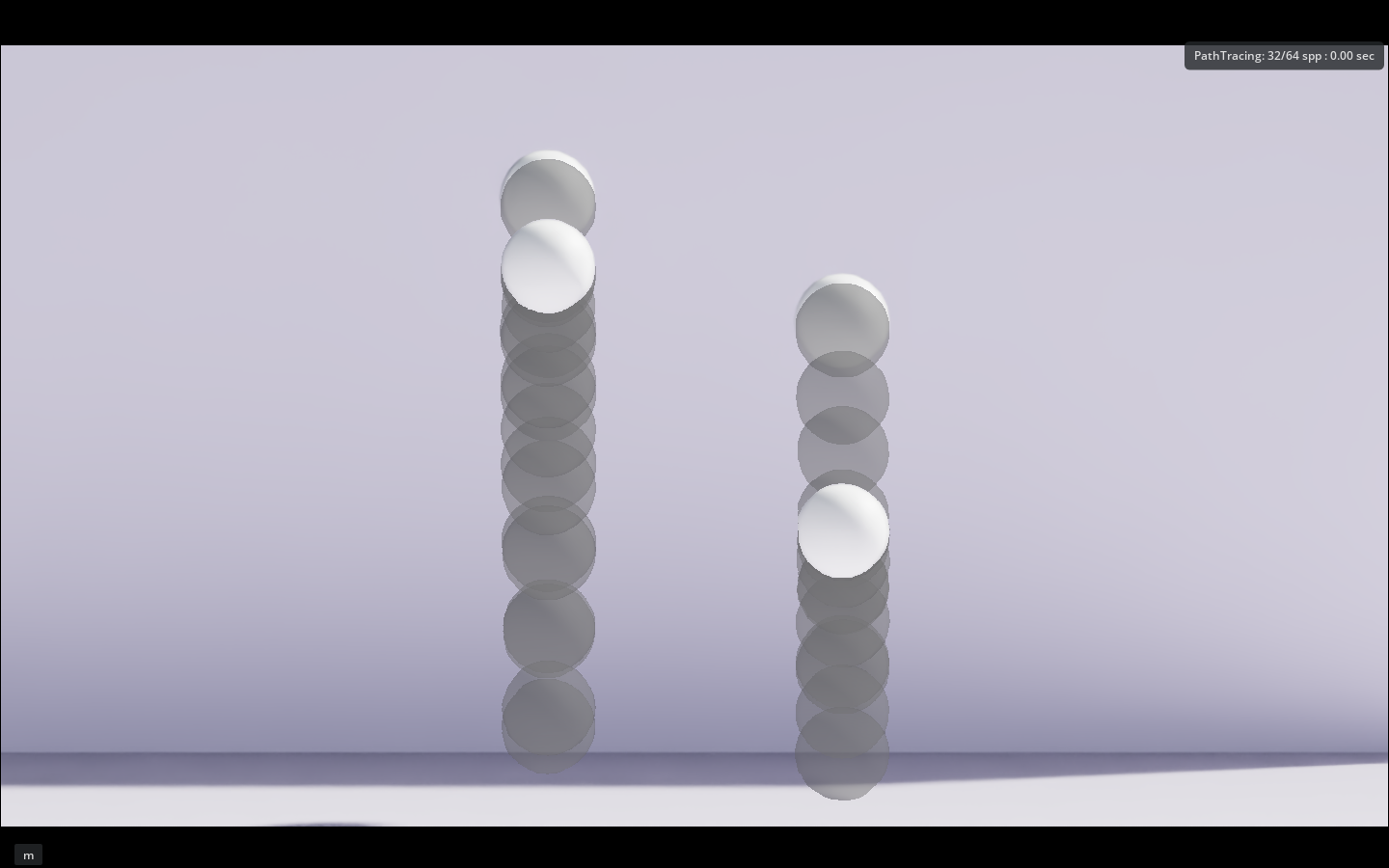}
    \caption{Restitution}
\end{subfigure}\hfill
\begin{subfigure}[t]{0.24\linewidth}
    \centering
    \includegraphics[
    width=\linewidth,
    trim=0px 50px 0px 130px,
    clip
]{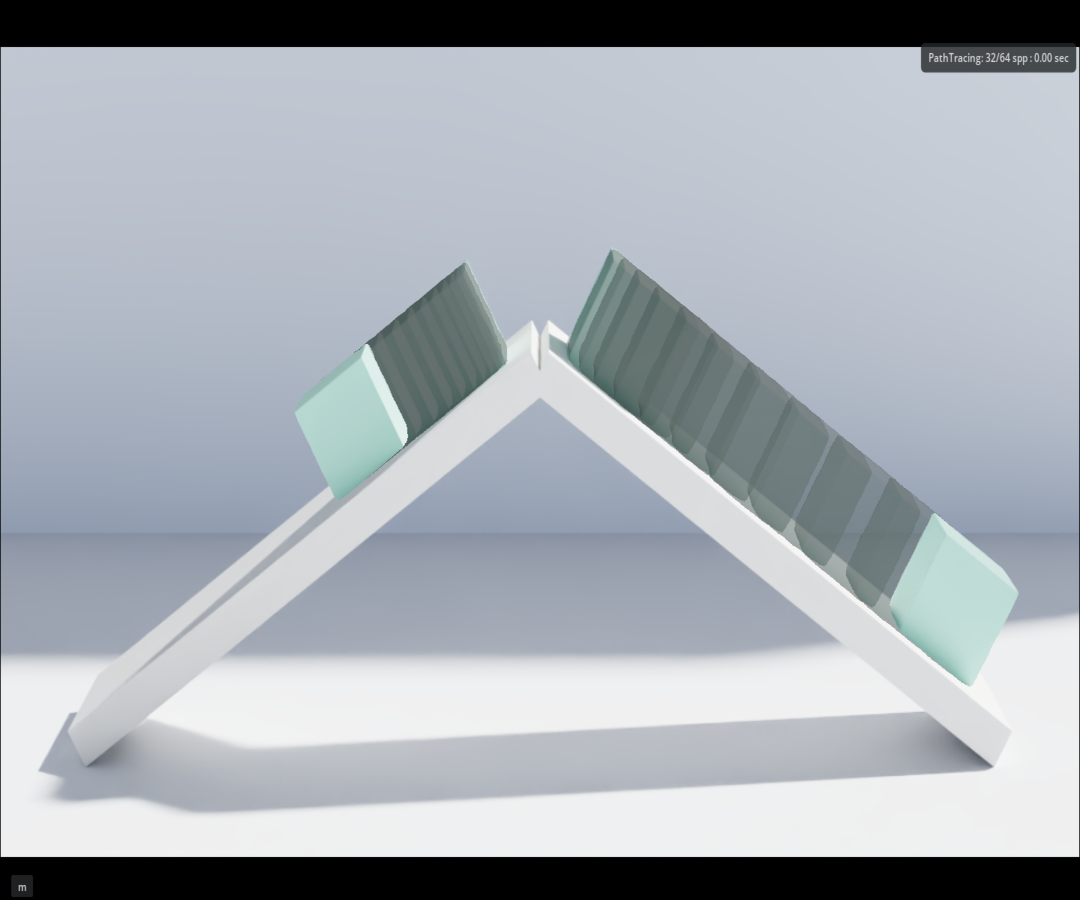}
    \caption{Friction}
\end{subfigure}\hfill
\begin{subfigure}[t]{0.24\linewidth}
    \centering
    \includegraphics[
    width=\linewidth,
    trim=180px 50px 180px 130px,
    clip
]{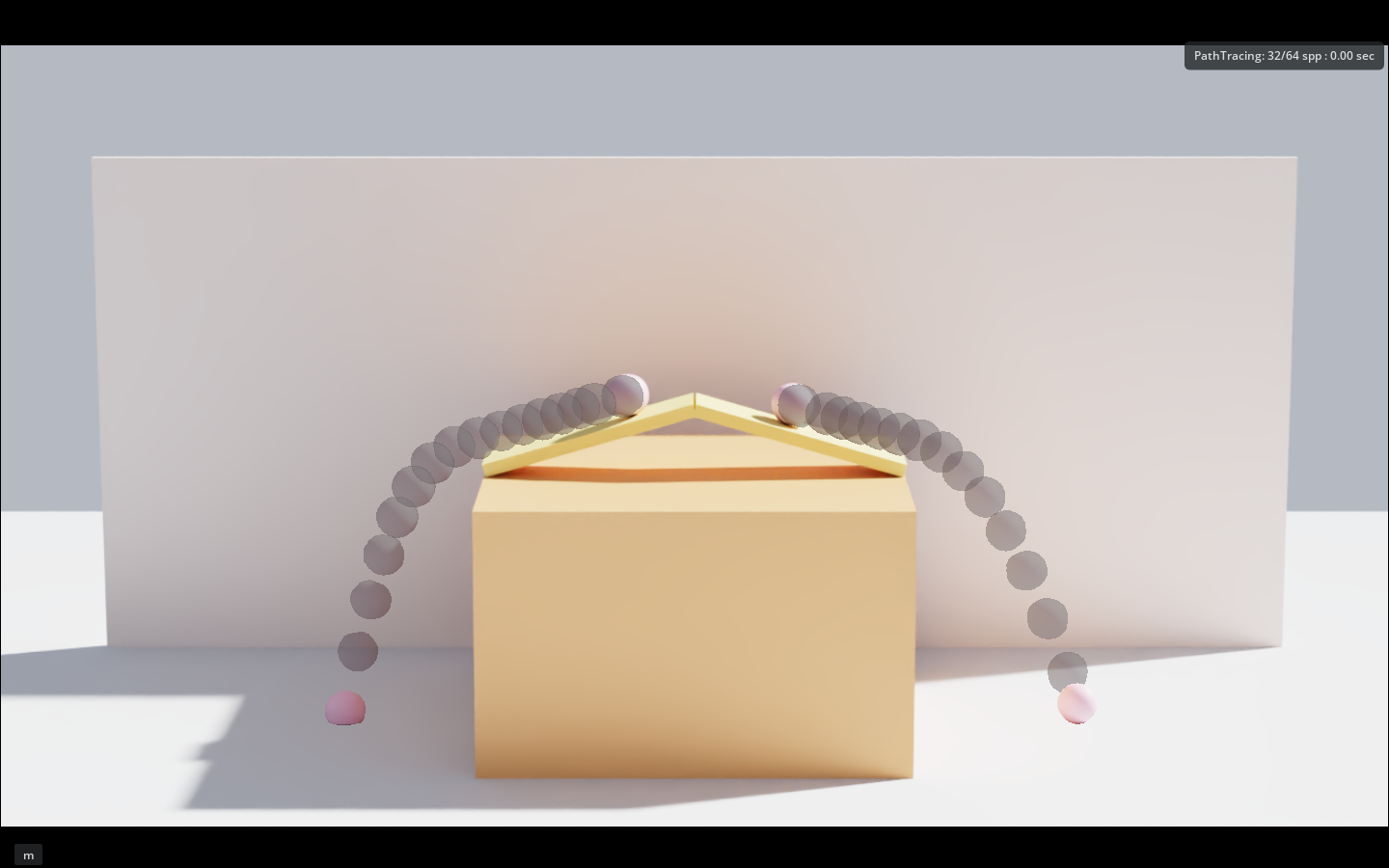}
    \caption{Projectile}
\end{subfigure}\hfill
\begin{subfigure}[t]{0.24\linewidth}
    \centering
    \includegraphics[
    width=\linewidth,
    trim=180px 50px 180px 130px,
    clip
]{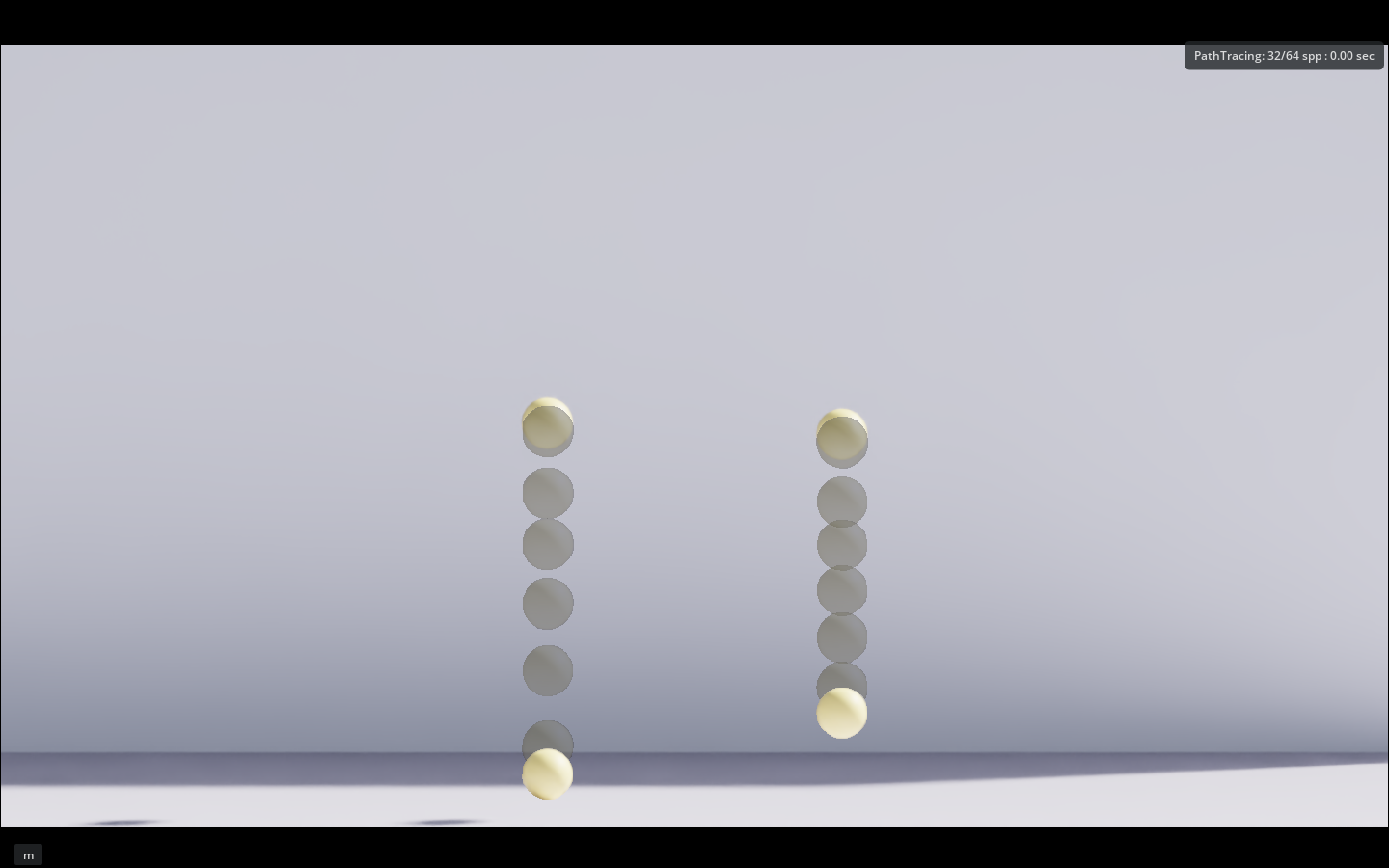}
    \caption{Gravity}
\end{subfigure}

\vspace{2mm}

\begin{subfigure}[t]{0.24\linewidth}
    \centering
    \includegraphics[
    width=\linewidth,
    trim=0px 50px 0px 130px,
    clip
]{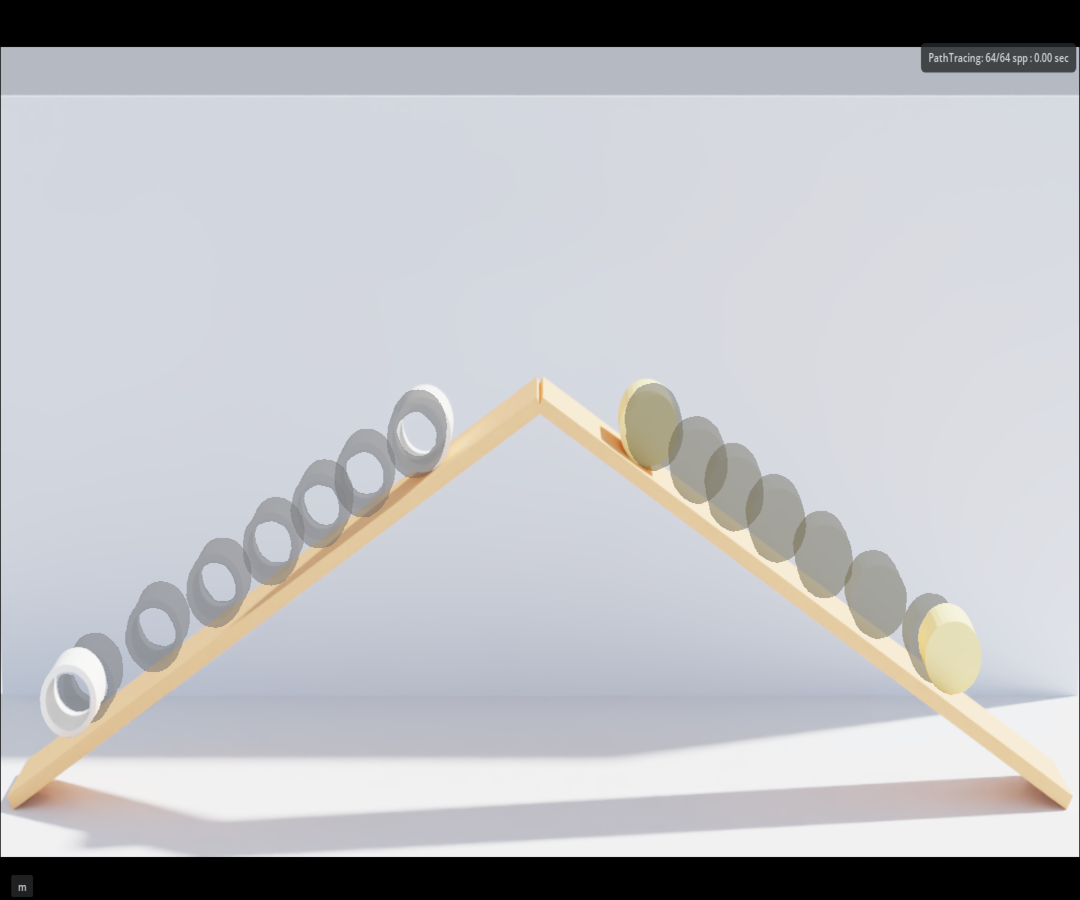}
    \caption{Rotational Inertia}
\end{subfigure}\hfill
\begin{subfigure}[t]{0.24\linewidth}
    \centering
    \includegraphics[
    width=\linewidth,
    trim=180px 50px 180px 130px,
    clip
]{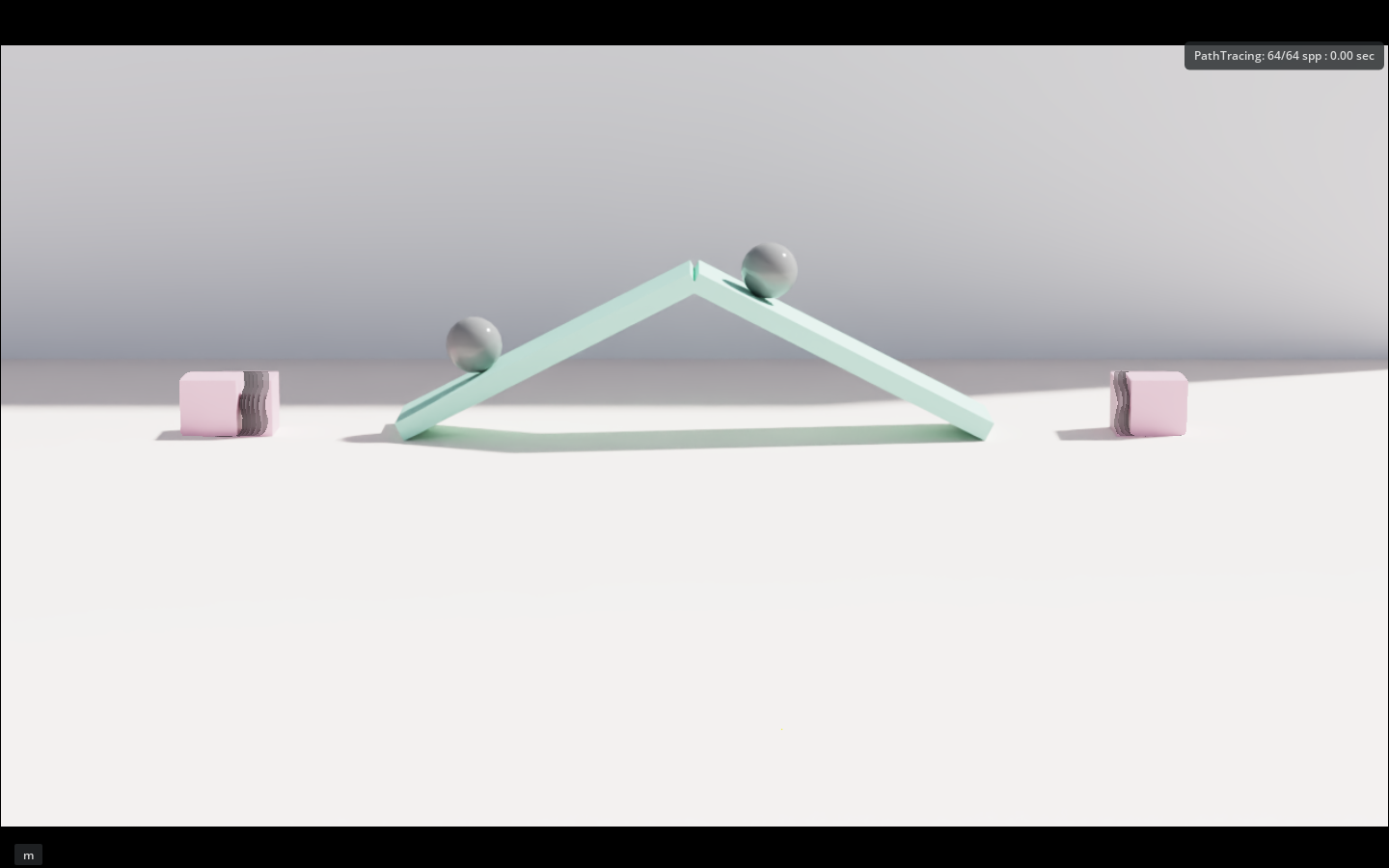}
    \caption{Momentum}
\end{subfigure}\hfill
\begin{subfigure}[t]{0.24\linewidth}
    \centering
    \includegraphics[
    width=\linewidth,
    trim=0px 50px 0px 130px,
    clip
]{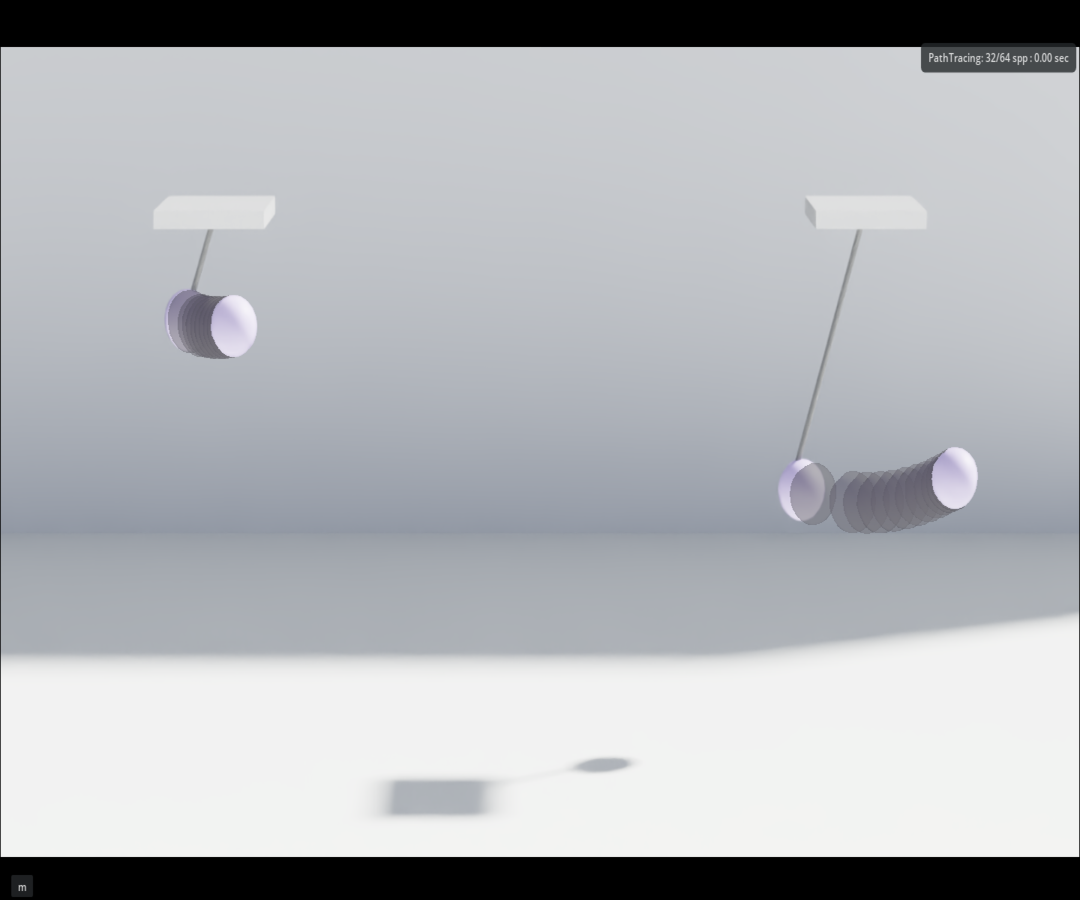}
    \caption{Pendulum}
\end{subfigure}\hfill
\begin{subfigure}[t]{0.24\linewidth}
    \centering
    \includegraphics[
    width=\linewidth,
    trim=180px 90px 180px 90px,
    clip
]{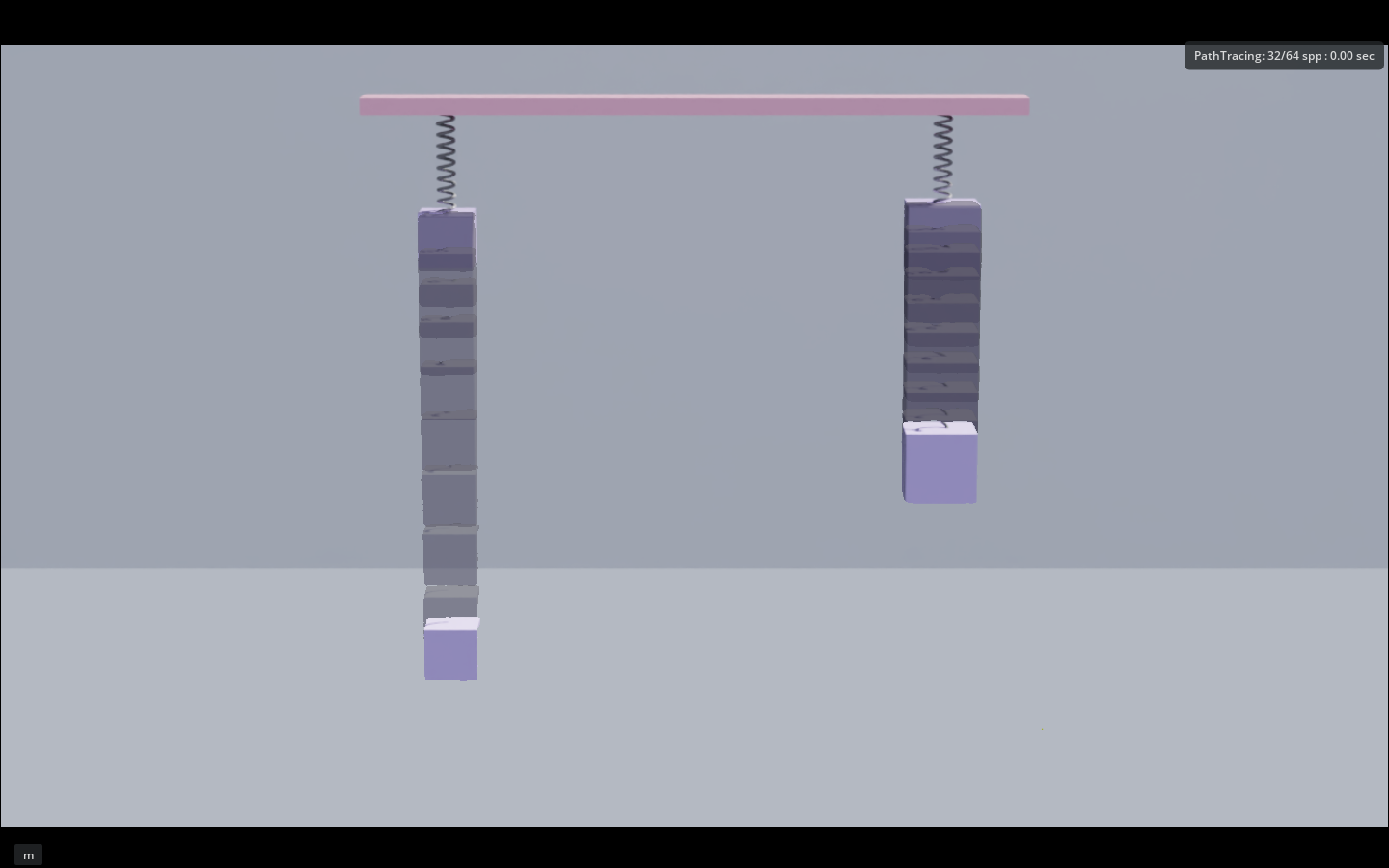}
    \caption{Spring}
\end{subfigure}

\caption{\textbf{Representative anti-physics examples from Principia-Synth.} Each scene intentionally violates the relational invariant of the corresponding physical phenomenon. From left to right, top to bottom: restitution (inconsistent restitution), friction (mass-dependent sliding despite identical surface properties), projectile motion (incorrect range ordering), gravity (incorrect fall-time relationship), rotational inertia (hollow cylinder reaches the bottom before the solid cylinder), momentum (a ball with greater initial momentum transfers less momentum to the corresponding block), pendulum (the longer pendulum oscillates with a shorter period), and spring (the lighter mass exhibits a larger equilibrium displacement than the heavier mass).}
\label{fig:antiphysics_examples}
\end{figure}

We use Omniverse\cite{NVIDIA_Isaac_Sim} to create synthetic videos having relational physics violations creating an \textit{anti-physics} set. Representative examples for each scenario are in Fig.\ref{fig:antiphysics_examples}.
\label{app:principia-synth-anti-physics}

\subsection{Evaluation Protocol}
\label{app:vlm_protocol}

For each scene, we present the VLM with the full real world physics or anti-physics video and a phenomenon-specific prompt asking the model to classify the physics as PASS or FAIL with a brief explanation. The exact prompts used are:

\begin{itemize}[leftmargin=*]

\item \textbf{Restitution:} \textit{Check bouncing physics of the two objects. Both objects are identical. Limit the analysis to the first bounce only. Return ONLY valid JSON (no markdown, no code fences) in the following format: \texttt{\{"verdict":"PASS"|"FAIL","reason":"brief explanation"\}}.}

\item \textbf{Gravity:} \textit{Check free fall physics of the two objects, specifically the time taken to hit the ground. Both objects are identical. Limit the analysis to the initial fall only, before any bouncing occurs. Return ONLY valid JSON (no markdown, no code fences) in the following format: \texttt{\{"verdict":"PASS"|"FAIL","reason":"brief explanation"\}}.}

\item \textbf{Friction:} \textit{Check sliding physics of the two blocks. Both are made of the same material. Return ONLY valid JSON (no markdown, no code fences) in the following format: \texttt{\{"verdict":"PASS"|"FAIL","reason":"brief explanation"\}}.}

\item \textbf{Rotational Inertia:} \textit{Check rolling dynamics of the hollow and solid cylinder. Both have the same mass and radius. Return ONLY valid JSON (no markdown, no code fences) in the following format: \texttt{\{"verdict":"PASS"|"FAIL","reason":"brief explanation"\}}.}

\item \textbf{Projectile:} \textit{Check projectile motion trajectory of the two balls, specifically the horizontal range. Both balls are identical. Return ONLY valid JSON (no markdown, no code fences) in the following format: \texttt{\{"verdict":"PASS"|"FAIL","reason":"brief explanation"\}}.}

\item \textbf{Momentum:} \textit{Check collision physics of the balls with the blocks, specifically the distance travelled by the blocks after the collision. Both balls have the same mass. The blocks are solid and are made of the same material. Return ONLY valid JSON (no markdown, no code fences) in the following format: \texttt{\{"verdict":"PASS"|"FAIL","reason":"brief explanation"\}}.}

\item \textbf{Mass-Spring:} \textit{Check extension behaviour of the springs when mass is placed on them. Both springs have the same spring constant. Both blocks are solid and are made of the same material. Return ONLY valid JSON (no markdown, no code fences) in the following format: \texttt{\{"verdict":"PASS"|"FAIL","reason":"brief explanation"\}}.}

\item \textbf{Pendulum:} \textit{Check pendulum motion physics, specifically the time period of the swings. Return ONLY valid JSON (no markdown, no code fences) in the following format: \texttt{\{"verdict":"PASS"|"FAIL","reason":"brief explanation"\}}.}

\end{itemize}

We score VLM responses by agreement with the ground-truth Principia-synth binary score: a VLM scene receives credit if its PASS/FAIL judgment matches whether the video actually satisfies the relational invariant. The agreement score $A_\phi \in [0, 1]$ is the fraction of scenes in phenomenon $\phi$ where the VLM and the Principia-synth ground truth agree.

\begin{figure}[t]
\centering
\includegraphics[width=\linewidth]{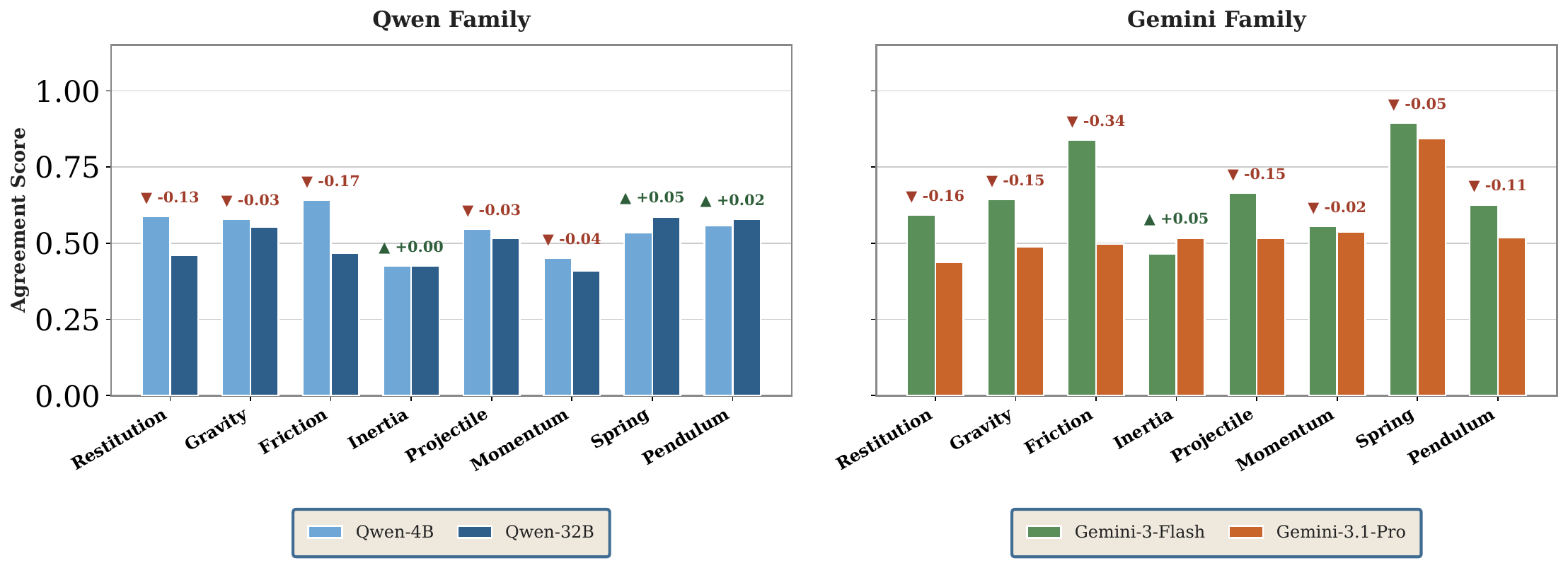}
\caption{\textbf{VLM scaling effects.} 
Per-phenomenon agreement changes from Qwen-4B to Qwen-32B (left, $8\times$ scale) and Gemini-3-Flash to Gemini-3.1-Pro (right). 
Scaling does not lead to an increase in performance.}
\label{fig:vlm_scaling}
\end{figure} 

\subsection{VLM Scaling}

Figure~\ref{fig:vlm_scaling} compares within-family scaling for Qwen and Gemini. Scaling does not improve performance, with Gemini and Qwen showing overall performance reductions of $0.12$ and $0.04$, respectively. For both model families, Friction exhibits the largest regression, with decreases of $0.34$ for Gemini and $0.17$ for Qwen.

\section{Additional Qualitative Results of Video Generators on Principia}
\begin{figure*}[t!]
  \centering
\includegraphics[width=\linewidth,keepaspectratio]{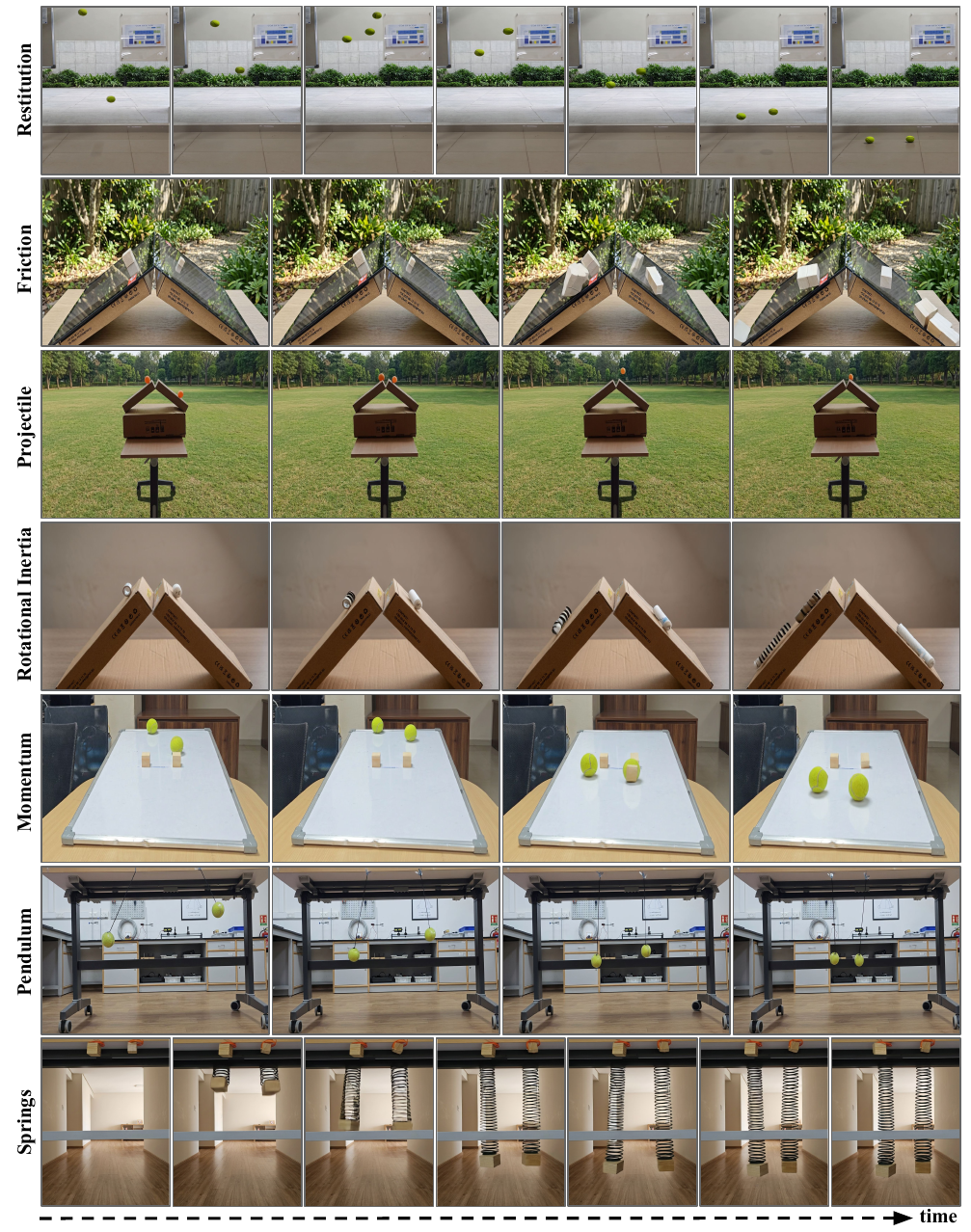}
    \caption{\textbf{Qualitative results of Wan2.2-5B}. 
    \textbf{(1) Restitution:} Two identical balls are dropped from different heights. The generated video fails to produce plausible falling motion and hallucinates a third ball in the scene.
    \textbf{(2) Friction:} The model produces implausible sliding motion of the blocks, which additionally hallucinate and deform during motion.
    \textbf{(3) Projectile:} The generated video fails to produce plausible projectile motion -- both balls move upward, and the left ball disappears during the sequence.
    \textbf{(4) Inertia:} The generated video produces incorrect rolling motion of the cylinders, which additionally deform into different objects.
    \textbf{(5) Momentum:} The generated video produces implausible motion in the blocks following the collisions.
    \textbf{(6) Pendulum:} The generated video fails to produce correct oscillatory motion in the bobs -- both bobs swing toward the center and then hover.
    \textbf{(7) Springs:} The generated video produces implausible motion -- both blocks travel similar distances despite having different masses, and both springs hallucinate into different colors and shapes.}
    \label{fig:supple-qual-wan5b}
    \vspace{-10pt}
\end{figure*}

\begin{figure*}[t!]
  \centering
\includegraphics[width=\linewidth,keepaspectratio]{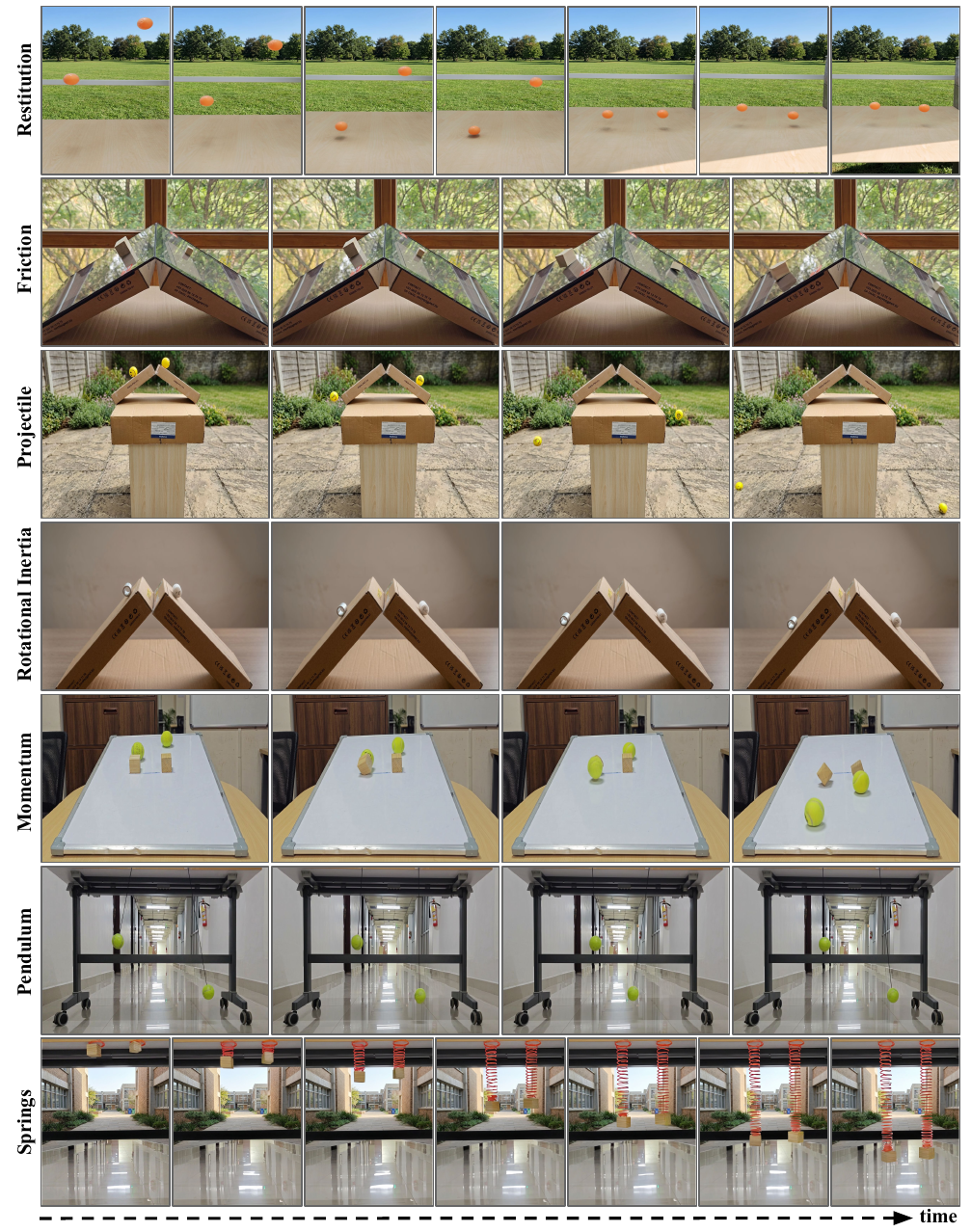}
    \caption{\textbf{Qualitative results of Wan2.2-14B}. 
    \textbf{(1) Restitution:} Two identical balls are dropped from different heights. The higher ball fails to rebound to a proportionally higher height than the lower ball, violating the expected restitution relationship.
    \textbf{(2) Friction:} The model produces implausible sliding motion of the blocks, which additionally hallucinate and deform during motion.
    \textbf{(3) Projectile:} The ball launched from the higher height fails to travel a larger horizontal distance than the other ball, violating the expected projectile-motion relationship.
    \textbf{(4) Inertia:} The generated video produces incorrect rolling motion of the cylinders -- the solid cylinder travels approximately the same distance as the hollow cylinder.
    \textbf{(5) Momentum:} The generated video produces implausible motion in the blocks following the collisions.
    \textbf{(6) Pendulum:} The generated video fails to produce oscillatory motion in the bobs -- both bobs swing toward the center and then hover.
    \textbf{(7) Springs:} The generated video produces implausible motion -- both blocks travel similar distances despite having different masses.}
    \label{fig:supple-qual-wan14b}
    \vspace{-10pt}
\end{figure*}

\begin{figure*}[t!]
  \centering
\includegraphics[width=\linewidth,keepaspectratio]{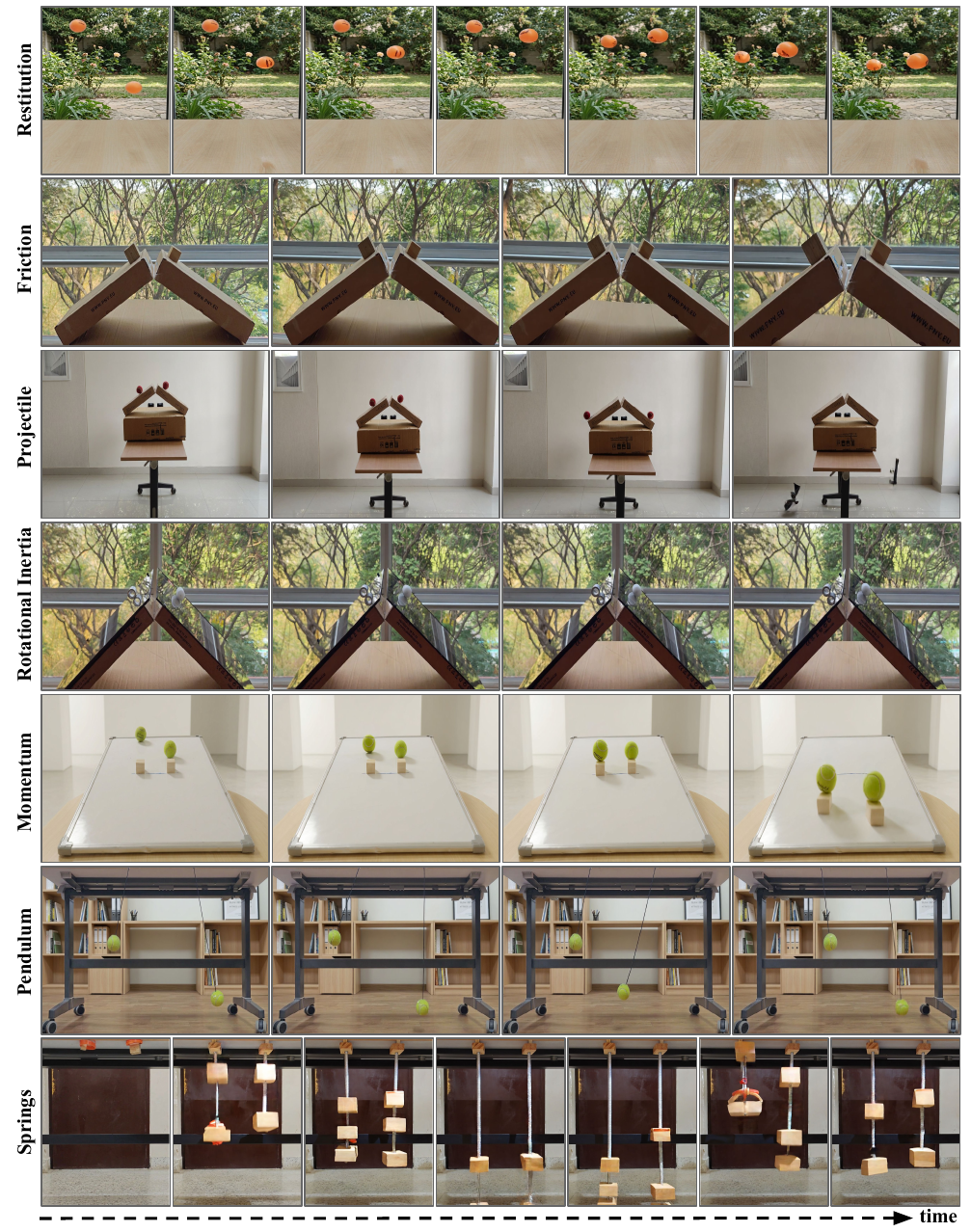}
    \caption{\textbf{Qualitative results of Cosmos2.5-2B}. 
    \textbf{(1) Restitution:} Two identical balls are dropped from different heights. Both balls hover in the air and fail to fall to the ground.
    \textbf{(2) Friction:} The model fails to produce sliding motion in the blocks, and both blocks remain stationary.
    \textbf{(3) Projectile:} The balls launch toward the ground but hallucinate into different objects during motion.
    \textbf{(4) Inertia:} The generated video fails to produce any rolling motion in the cylinders, and both remain stationary.
    \textbf{(5) Momentum:} The balls slide down and collide with the blocks before jumping onto them. Although the blocks begin to slide, the traveled distances are not proportional to the transferred momentum -- the ball released from the higher height fails to move its corresponding block farther.
    \textbf{(6) Pendulum:} The generated video produces implausible motion in the bobs -- the left bob hovers momentarily while the right bob swings to the opposite side.
    \textbf{(7) Springs:} The generated video produces implausible motion -- both springs and both blocks hallucinate into multiple different objects.}
    \label{fig:supple-qual-cosmos2b}
    \vspace{-10pt}
\end{figure*}

\begin{figure*}[t!]
  \centering
\includegraphics[width=\linewidth,keepaspectratio]{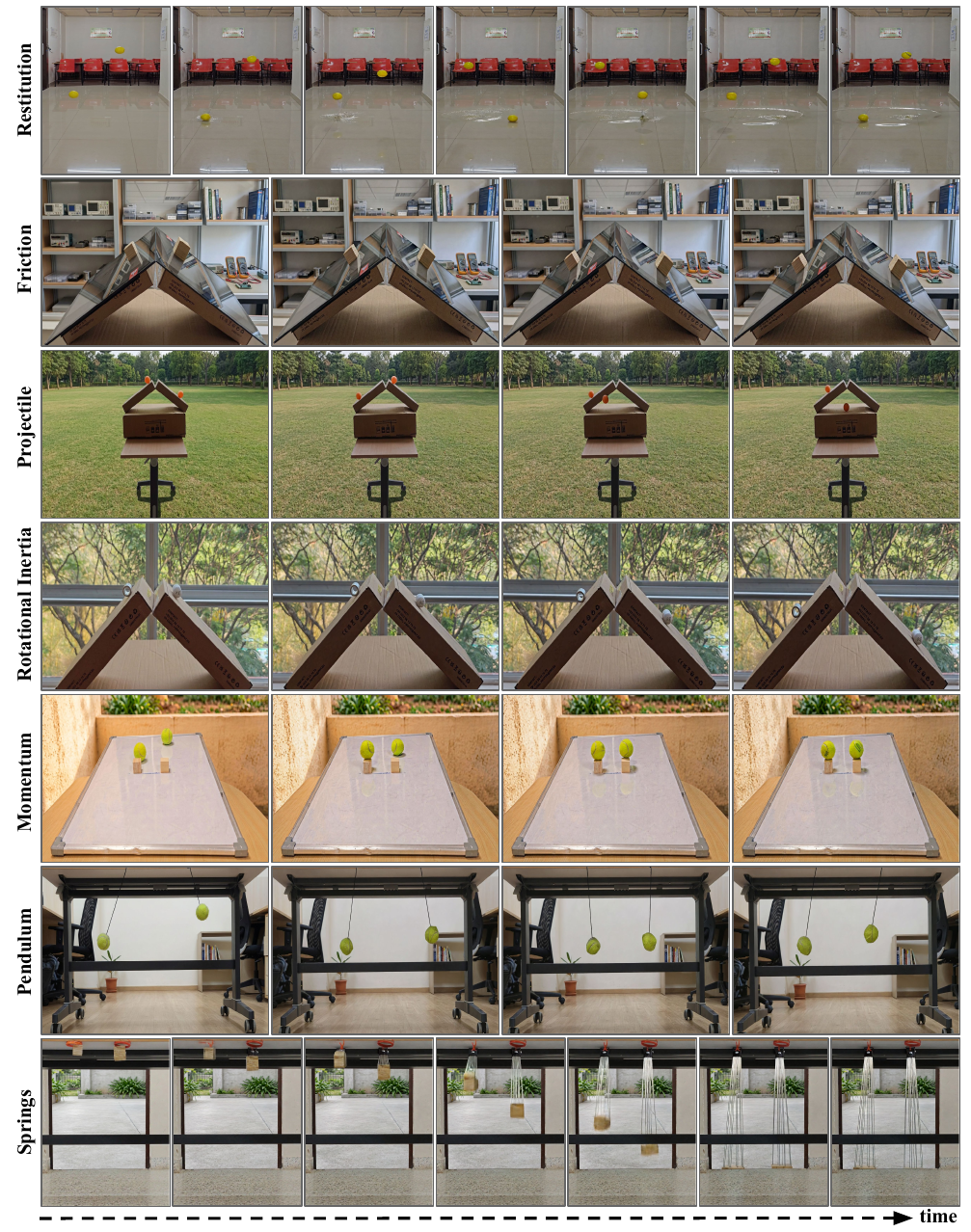}
    \caption{\textbf{Qualitative results of Cosmos2.5-14B}. 
    \textbf{(1) Restitution:} Two identical balls are dropped from different heights. The higher ball rebounds to a proportionally higher height than the lower ball, producing physically plausible behavior.
    \textbf{(2) Friction:} The model produces plausible sliding motion of both the blocks, and preserves the mass-independent timing.
    \textbf{(3) Projectile:} The right ball fails to follow a plausible projectile trajectory, instead landing on the cardboard box rather than the ground.
    \textbf{(4) Inertia:} The generated video produces correct rolling motion of both the cylinders -- the solid cylinder reaches before the hollow cylinder.
    \textbf{(5) Momentum:} The generated video produces implausible motion -- both balls jump onto the blocks and fail to induce any meaningful motion in the blocks.
    \textbf{(6) Pendulum:} The generated video produces implausible motion in the bobs -- both bobs have similar time periods despite having different lengths.
    \textbf{(7) Springs:} The generated video produces plausible motion -- the heavier block stretches the spring further downward, although the spring visuals hallucinate into a different object.}
    \label{fig:supple-qual-cosmos14b}
    \vspace{-10pt}
\end{figure*}

\begin{figure*}[t!]
  \centering
\includegraphics[width=\linewidth,keepaspectratio]{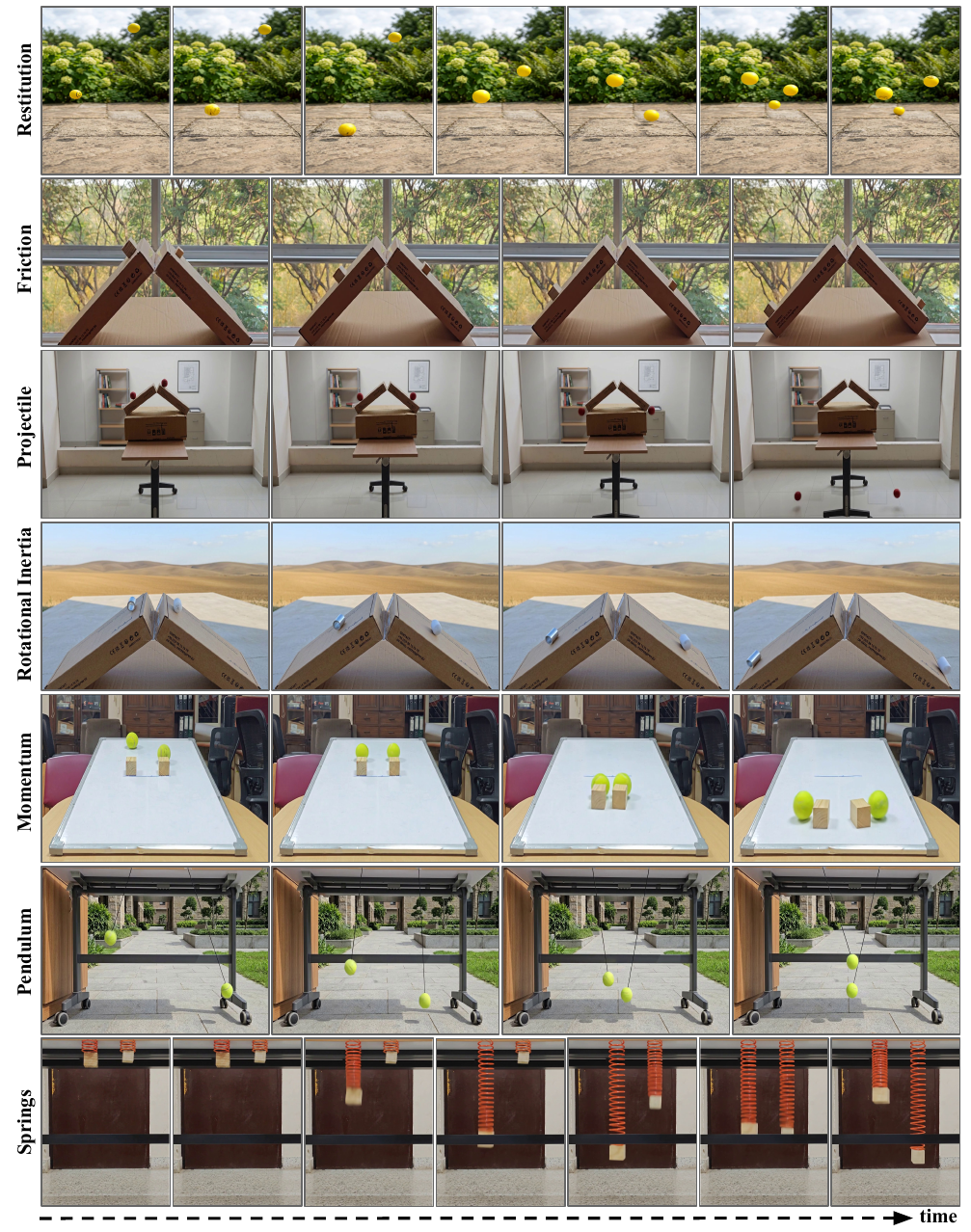}
    \caption{\textbf{Qualitative results of Veo-3.1}. 
    \textbf{(1) Restitution:} Two identical balls are dropped from different heights. The higher ball fails to rebound to a proportionally similar height as the lower ball, violating the expected restitution relationship.
    \textbf{(2) Friction:} The model produces plausible sliding motion of both the blocks, and preserves the mass-independent timing.
    \textbf{(3) Projectile:} The ball launched from a higher height fails to travel a larger horizontal distance than the other ball, violating the expected projectile-motion relationship.
    \textbf{(4) Inertia:} The generated video produces correct rolling motion of both the cylinders -- the solid cylinder reaches before the hollow cylinder
    \textbf{(5) Momentum:} The generated video produces implausible motion in the blocks -- both blocks travel identical distances.
    \textbf{(6) Pendulum:} The generated video produces implausible motion in the bobs-- both bobs have similar time periods despite having different lengths.
    \textbf{(7) Springs:} The generated video produces implausible motion in the blocks -- both blocks travel similar distances despite having different mass.}
    \label{fig:supple-qual-veo}
    \vspace{-10pt}
\end{figure*}

We provide model-wise results on our tasks in Figures\ref{fig:supple-qual-wan5b} - \ref{fig:supple-qual-veo}. We also recommend viewing the videos on the project webpage: https://principiabench.github.io/. 

\begin{figure*}[t!]
  \centering
\includegraphics[width=\linewidth,keepaspectratio]{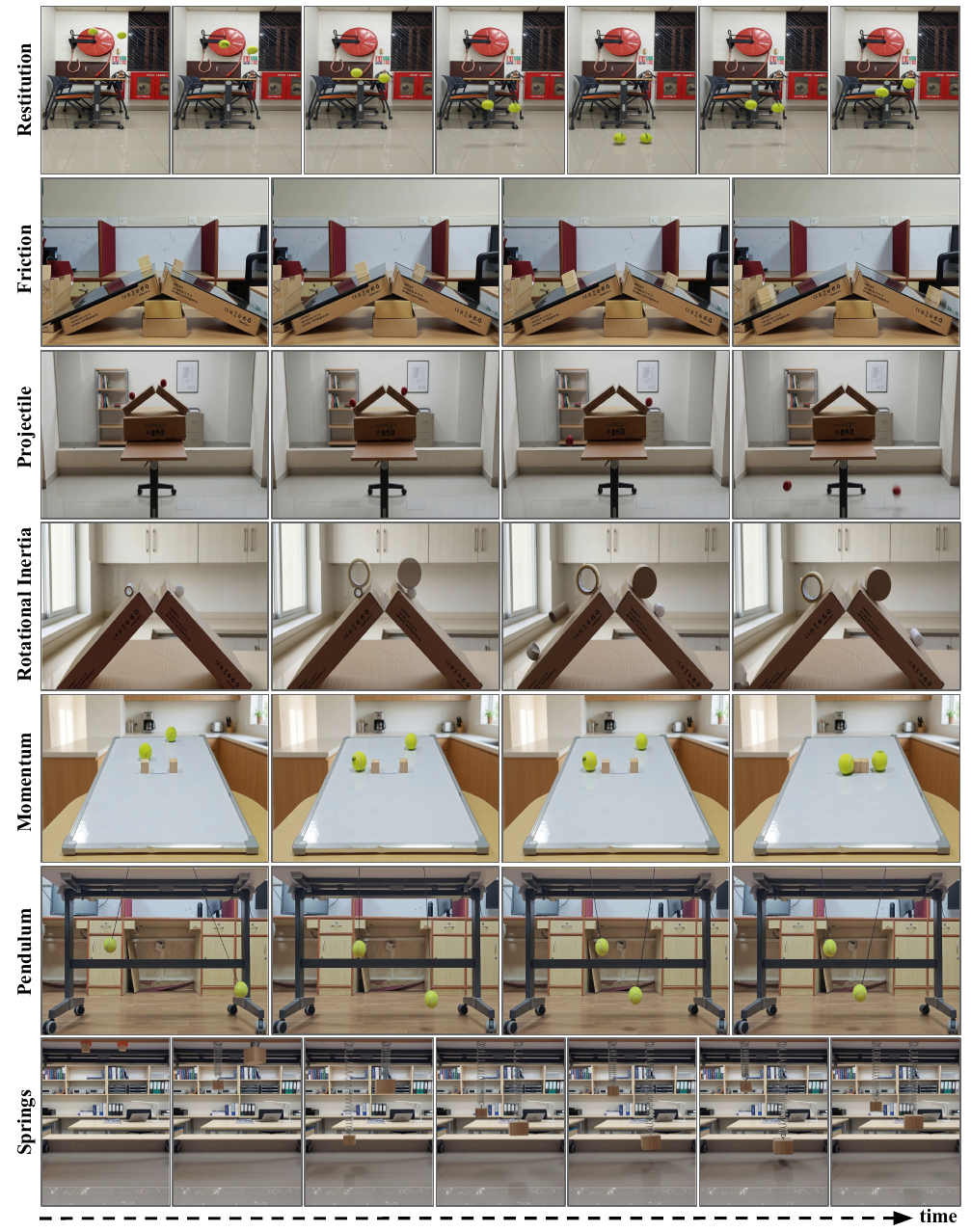}
    \caption{\textbf{Qualitative results of Omni}. 
    \textbf{(1) Restitution:} Two identical balls are dropped from different heights. The higher ball fails to rebound to a proportionally similar height as the lower ball, violating the expected restitution relationship.
    \textbf{(2) Friction:} The model produces plausible sliding motion of both the blocks, and preserves the mass-independent timing.
    \textbf{(3) Projectile:} The ball launched from a higher height fails to travel a larger horizontal distance than the other ball, violating the expected projectile-motion relationship.
    \textbf{(4) Inertia:} The generated video fails to produce correct rolling motion of both the cylinders -- the hollow cylinder reaches before the solid cylinder and duplicate cylinders appear at the top of the inclines.
    \textbf{(5) Momentum:} The generated video produces implausible motion in the blocks.
    \textbf{(6) Pendulum:} The generated video produces implausible motion in the bobs -- both bobs have similar time periods despite having different lengths.
    \textbf{(7) Springs:} The generated video produces implausible motion in the blocks -- the right block hallucinates to a larger size and drops further than the left block.}
    \label{fig:supple-qual-omni}
    \vspace{-10pt}
\end{figure*}

\end{document}